\documentclass[a4paper,fleqn]{cas-dc}

\usepackage[numbers]{natbib} %

\def\tsc#1{\csdef{#1}{\textsc{\lowercase{#1}}\xspace}}
\tsc{WGM}
\tsc{QE}

\RenewDocumentCommand \printorcid { } { }

\newproof{proof}{Proof} %

\usepackage{amsmath,amssymb,amsfonts}
\usepackage{graphicx}
\usepackage{algorithm,algorithmic}

\usepackage{booktabs}
\usepackage{multirow}
\usepackage{threeparttable}
\newtheorem{assumption}{Assumption}
\newtheorem{lemma}{Lemma}

\newtheorem{theorem}{Theorem}

\usepackage{xcolor}
\usepackage[normalem]{ulem}

\newif\ifshowchanges

\showchangesfalse

\newcommand{\rev}[1]{%
  \ifshowchanges
    {\color{blue}#1}%
  \else
    #1%
  \fi
}

\newcommand{\del}[1]{%
  \ifshowchanges
    {\color{red}\sout{#1}}%
  \else
  \fi
}

\newif\ifcomments\commentsfalse

\newcommand{\LL}[1]{
    \ifcomments
        \textcolor{blue}{LL: #1}
    \else
    \fi
    }

\begin{document}
\let\WriteBookmarks\relax
\def\floatpagepagefraction{1}
\def\textpagefraction{.001}

\shorttitle{}    

\shortauthors{L.~L\"uken, S.~Lucia}  

\title [mode = title]{Self-Supervised Learning of Iterative Solvers for Constrained Optimization}

\author[1]{Lukas~L\"uken}%
\cormark[1]
\credit{Writing - original draft, Investigation, Visualization, Software, Methodology, Formal analysis, Conceptualization}
\ead[1]{lukas.lueken@tu-dortmund.de}
\cortext[1]{Corresponding author}

\author[1]{Sergio~Lucia}%
\credit{Writing - review and editing, Supervision, Conceptualization}
\ead[2]{sergio.lucia@tu-dortmund.de}

\affiliation[1]{organization={Chair of Process Automation Systems, TU Dortmund University},
            addressline={Emil-Figge Straße 70}, 
            city={Dortmund},
            postcode={44227}, 
            state={North Rhine-Westphalia},
            country={Germany}}

\nonumnote{This is the author accepted manuscript of the article published in Results in Control and Optimization. The final published version is available at: \url{https://doi.org/10.1016/j.rico.2026.100751}. This manuscript version is made available under the CC BY-NC-ND 4.0 license.}

\begin{abstract}
The real-time solution of parametric optimization problems is critical for applications that demand high accuracy under tight real-time constraints, such as model predictive control.
To this end, this work presents a learning-based iterative solver for constrained optimization, comprising a neural network predictor that generates initial primal-dual solution estimates, followed by a learned iterative solver that refines these estimates to reach high accuracy.

We introduce a novel loss function based on Karush-Kuhn-Tucker (KKT) optimality conditions, enabling fully self-supervised training without pre-solved optimizer solutions.
Theoretical guarantees ensure that the training loss function attains minima exclusively at KKT points.
A convexification procedure enables application to nonconvex problems while preserving these guarantees.

Experiments on two nonconvex case studies demonstrate speedups of up to one order of magnitude compared to state-of-the-art solvers such as IPOPT, while achieving orders of magnitude higher accuracy than competing learning-based approaches. 

\end{abstract}


\maketitle

\section{Introduction}
\label{sec:introduction}

The accurate and fast solution of constrained optimization problems as a function of their parameters is of great importance for a variety of applications, such as control \cite{arangoNeuralNetworksFast2023}, optimal power flow \cite{dontiDC3LearningMethod2021, parkSelfSupervisedPrimalDualLearning2023}, or planning and scheduling \cite{kotaryFastApproximationsJob2022}. 
However, real-time requirements pose a major challenge, especially when accurate solutions are required.

This challenge becomes particularly evident in the context of model predictive control (MPC), where a finite-horizon optimal control problem is solved at each sampling instant with varying parameters, such as the current state of the system. 
The first computed control input is applied and the process is repeated in a receding-horizon fashion.
MPC naturally handles hard input and state constraints and can consider complex, multi-objective performance criteria.
MPC is in principle well suited to control complex nonlinear systems, but its real-time implementation poses significant challenges as the computational burden grows rapidly with model complexity, horizon length and number of constraints.
Therefore, its application is limited for systems with fast dynamics and very tight hardware constraints or in the case of complex nonlinear MPC problems such as in robust NMPC formulations or large-scale systems \cite{kumarIndustrialLargescaleModel2021,kargReinforcedApproximateRobust2021}.
\LL{find better references to support claim of this paragraph, see arango paper for references.}

To address this challenge, various approaches have been proposed in the context of MPC and constrained optimization in general.

Explicit MPC \cite{bemporadExplicitLinearQuadratic2002} pre-solves the mapping from state to optimal input offline, storing it as a piecewise‐defined control law.
While efficient online, explicit schemes suffer from exponential memory growth and are typically only viable for small, linear or mildly nonlinear problems.

Neural network (NN) approximations of MPC \cite{parisiniRecedinghorizonRegulatorNonlinear1995,arangoNeuralNetworksFast2023,hertneckLearningApproximateModel2018,kargEfficientRepresentationApproximation2020} or NN approximations of constrained optimization in general \cite{dontiDC3LearningMethod2021,nikbakhtUnsupervisedLearningParametric2021,amosTutorialAmortizedOptimization2023,parkSelfSupervisedPrimalDualLearning2023} have been used to replace online optimization with fast feedforward neural network evaluations.
Recent advances leveraging deep neural networks achieve multiple orders of magnitude reduction in computation time for large, highly nonlinear MPC problems \cite{kargReinforcedApproximateRobust2021,kumarIndustrialLargescaleModel2021} while lowering the memory footprint significantly \cite{bonzaniniSafeDoseDelivery2020}, making neural network approximations attractive for embedded deployment \cite{luciaDeepLearningBasedModel2021}\del{.} \rev{ and for safety-critical engineering applications such as physical human-robot interaction \cite{nurbayevaDeepImitationLearning2023}.}

Despite their speed, learning-based approximations of the parametric optimization problem solution face important challenges.
First, large amounts of training data are required and sampling this data by solving many optimization instances offline can be prohibitively expensive \cite{lukenSobolevTrainingDataefficient2023,krishnamoorthyImprovedDataAugmentation2023}.
Second, the neural network predictions are subject to approximation errors, which may lead to constraint violations or suboptimal inputs, undermining safety and performance.
To ensure constraint satisfaction and a safe application of the neural network, additional measures have to be taken such as probabilistic validation \cite{hertneckLearningApproximateModel2018,kargProbabilisticPerformanceValidation2021,nubertSafeFastTracking2020}, deterministic verification \cite{adamekDeterministicSafetyGuarantees2024,fabianiReliablyStabilizingPiecewiseAffineNeural2023,teichribErrorBoundsMaxout2023} or projection onto feasible sets \cite{kargEfficientRepresentationApproximation2020,chenApproximatingExplicitModel2018}.
However, these often lead to suboptimal performance or significantly increase the computational burden, which is counterproductive to the goal of fast online optimization.

Recently, learning-based approximation approaches for parametric constrained optimization problems have been proposed outside of the MPC context.
For example, neural networks have been used to predict optimal solutions for optimal power flow \cite{dontiDC3LearningMethod2021} or job shop scheduling \cite{kotaryFastApproximationsJob2022}.
These approaches though suffer from similar issues as MPC approximations, such as approximation errors leading to constraint violations and suboptimal solutions.
To mitigate these issues, methods like DC3 \cite{dontiDC3LearningMethod2021} augment the neural network predictions with Newton-like correction steps to enforce feasibility.
This approach, however, significantly increases the runtime and only provides suboptimal solutions.
The Primal-Dual Learning (PDL) method \cite{parkSelfSupervisedPrimalDualLearning2023} trains two networks in a self-supervised loop using augmented Lagrangian losses, eliminating reliance on pre-solved optimizer solutions.
PDL has been successfully applied to MPC problems by including a closed-loop training procedure, but still is prone to approximation errors and remaining optimality gaps \cite{frommeUnsupervisedClosedloopPrimaldual2025}.
\rev{The predict-then-optimize paradigm \cite{elmachtoub2022smart} uses the structure of an optimization problem to design prediction models for unknown parameters so that downstream decisions are of high quality.
Decision-focused learning \cite{mandiDecisionFocusedLearningFoundations2024} extends this idea to a broad class of methods combining machine learning with constrained optimization.
Unlike these approaches, LISCO assumes known parameters and learns to solve the resulting optimization problem itself, while providing a certifiable primal-dual solution.}

Recently, the field of learning-to-optimize (L2O) has emerged, which aims to learn optimization algorithms directly instead of approximating the solution to a parametric optimization problem.
As of today, L2O consists of a wide range of approaches, such as methods to learn neural network based warmstarts for classical optimization solvers \cite{chenLargeScaleModel2022,sambharyaEndtoEndLearningWarmStart2023} or to learn hyperparameters of classical optimization solvers \cite{sambharyaLearningAlgorithmHyperparameters2024}.
Also, approaches that directly predict approximate solutions to the parametric optimization problems, as previously described, have been labeled as L2O \cite{tangLearningOptimizeMixedInteger2024,kotaryLearningConstrainedOptimization2024}.
Further, methods have been proposed to learn neural network based optimizers for machine learning problems as a substitute for classical training algorithms such as stochastic gradient descent \cite{liLearningOptimize2016}.
A broad overview can be found in \cite{chenLearningOptimizePrimer2022}.
\LL{check consistency of references with statements and add more references}
\LL{better description of the ml l2o solvers. e.g. focus more on similarities to standard SGD}

However, most of these approaches do not address general nonlinear constrained optimization problems and their corresponding primal-dual solutions.
Therefore, no optimality certification can be provided.
Furthermore, current L2O methods target fast solutions of low accuracy for data-driven problems, such as neural network training (e.g., image classification), rather than high-accuracy solutions for parametric optimization problems.
Additionally, many approaches focus on learning the hyperparameters of classical solvers (e.g., step size and momentum) rather than the solvers themselves or the update steps.
While this may improve the performance of classical solvers, the solvers' fundamental limitations still apply.
Therefore, current L2O approaches do not fully address the requirements for solving specific parametric constrained optimization problems, such as MPC, which require high accuracy and certification of optimality.

We address the shortcomings of approximate MPC approaches, such as sampling costs and approximation errors, as well as the limitations of L2O methods, such as the lack of optimality certification and focus on low-accuracy solutions.
To this end, we propose a \emph{learning-based iterative solver for constrained optimization problems (LISCO)}.
This work builds upon our previous research introducing an initial version of LISCO \cite{lukenLearningIterativeSolvers2024}.
LISCO further extends the L2O paradigm and is specifically designed to obtain high-accuracy solutions for specific instances of parametric constrained optimization problems, such as nonlinear MPC.
\LL{add info on applicability. what type of problems are considered and how do they differ to L2O?}

The initial version \cite{lukenLearningIterativeSolvers2024} consists of a neural network that predicts update steps for a given current primal-dual iterate, analogous to Newton's method, iterating until convergence.
The network is trained self-supervised via residuals on the Karush-Kuhn-Tucker (KKT) conditions of optimality without requiring pre-solved optimizer solutions.
It achieves prediction accuracies multiple orders of magnitude better than established approximate MPC approaches, while providing a full primal-dual solution, which allows for certification of optimality, i.e. the KKT conditions are satisfied.

Despite the promising results, several limitations of the initial version of LISCO remain to be addressed.
A theoretical foundation of the KKT-based training loss function is missing, i.e. local minima of the loss function are not guaranteed to correspond to KKT points.
Furthermore, in the nonconvex case, these KKT points might correspond to maxima or saddle points.
Additionally, the initial version of LISCO places no specific focus on the initialization of the primal-dual iterates, which can lead to slow convergence and high runtimes.
Moreover, the original LISCO formulation inherently limits performance due to an overly restrictive formulation of the loss function based on linearized KKT residuals and requires the online calculation of second-order information for a line-search procedure.
The evaluation was limited to small problems without comparison to state-of-the-art learning-based approaches like DC3 or PDL \cite{lukenLearningIterativeSolvers2024}.

In this work, we propose an enhanced version of LISCO, incorporating several improvements to address the limitations of the original method.
The main contributions are:
\begin{enumerate}
    \item \textbf{Learning-based Iterative Solver for Constrained Optimization (LISCO):} We propose an extension to our previous work \cite{lukenLearningIterativeSolvers2024} and present LISCO, a neural network based iterative solver for constrained parametric optimization problems that predicts primal-dual update steps based on the current primal-dual iterate and the problem parameters.
    This approach is very fast to evaluate due to the evaluation speed of the neural network solver and achieves high-accuracy solutions, while providing a full primal-dual solution that allows for certification of optimality.
    \item \textbf{Improved Initialization of Primal-Dual Iterates:} We propose a neural network predictor that determines an initial primal-dual iterate based on the problem parameters.
    This improves the initialization of the primal-dual iterates, resulting in fewer iterations and lower runtimes.
    The predictor produces solutions that are comparable in accuracy to those of other learning-based approaches.
    \item \textbf{Self-Supervised Training Loss Function:} We propose a new training loss function based on KKT residuals of primal-dual iterates applicable to training the predictor and iterative solver networks.
    This function enables efficient self-supervised training of the networks without the need for pre-solved optimizer solutions and supports GPU acceleration.
    Additionally, we provide the theoretical foundation for the applicability of the KKT-based loss function for the self-supervised training of iterative solvers.
    We introduce a convexification procedure that enables training LISCO for nonconvex problems, such as nonlinear MPC.
    We remove the limitations of the original LISCO loss function, which was based on linearized KKT residuals and required an expensive line-search to train the iterative solver network.
\end{enumerate}

The remainder of this paper is structured as follows:
Section~\ref{sec:background} provides the mathematical background.
Section~\ref{sec:method} introduces LISCO, Section~\ref{sec:theoretical_properties} formalizes the applicability of the self-supervised loss function and Section~\ref{sec:implementation_details} describes the implementation details, which consider the convexification procedure as well as the training algorithm.
The method is applied to the nonlinear MPC of a nonlinear double integrator system as well as a larger nonconvex parametric optimization problem in Section~\ref{sec:case_study}.
Finally, Section~\ref{sec:conclusion} concludes the paper and discusses future work.\footnote{Code for this paper is available at: \url{https://github.com/lukaslueken/lisco-paper}}

\section{Background}
\label{sec:background}

We first formalize a general parametric nonlinear program (NLP) and its optimality conditions. 
We then describe the reformulation of the optimality conditions based on the Fischer-Burmeister complementary function \cite{fischerSpecialNewtontypeOptimization1992,chenPenalizedFischerBurmeisterNCPfunction2000}, which is fundamental for the learning-based iterative solver proposed in this work.
Finally, we briefly introduce the nonlinear model predictive control (NMPC) problem.

\subsection{Parametric Nonlinear Constrained Optimization}

A general parametric NLP can be described as follows:
\begin{subequations}\label{eq:nlp} 
    \begin{align}
        \min_{\mathbf{w}} \quad    &q\left(\mathbf{w},\mathbf{p}\right)\\
        \mathrm{s.t.} \quad &\mathbf{g}\left(\mathbf{w},\mathbf{p}\right) \leq \mathbf{0}, \\
        &\mathbf{h}\left(\mathbf{w}, \mathbf{p}\right) = \mathbf{0}.
    \end{align}
\end{subequations}

The goal is to minimize the objective function $q: \mathbb{R}^{n_w} \times \mathbb{R}^{n_p} \rightarrow \mathbb{R}$ with respect to the decision variables $\mathbf{w} \in \mathbb{R}^{n_w}$, given parameters $\mathbf{p} \in \mathbb{R}^{n_p}$, with respective dimensions $n_w$ and $n_p$. 
The optimization problem includes the inequality constraints $\mathbf{g}: \mathbb{R}^{n_w} \times \mathbb{R}^{n_p} \rightarrow \mathbb{R}^{n_g}$ and the equality constraints $\mathbf{h}: \mathbb{R}^{n_w} \times \mathbb{R}^{n_p} \rightarrow \mathbb{R}^{n_h}$.
These comprise $n_g$ individual inequality constraints $g_i$ for $i = 1, \ldots, n_g$ and $n_h$ individual equality constraints $h_j$ for $j = 1, \ldots, n_h$.
The objective function $q$ and the constraint functions $g_i$ and $h_j$ can be nonlinear and nonconvex, but are assumed to be smooth. 
Since the problem is parameterized by $\mathbf{p}$, the optimal solution $\mathbf{w}^*(\mathbf{p})$ is a function of the parameter vector $\mathbf{p}$.

A primal-dual solution of the optimization problem \eqref{eq:nlp} can be denoted by $\mathbf{z}^{*}(\mathbf{p}) = \left( \mathbf{w}^{*},\boldsymbol{\nu}^{*},\boldsymbol{\lambda}^{*} \right)$, with the Lagrange multipliers $\nu_j$ and $\lambda_i$ corresponding to the equality and inequality constraints $h_j$ and $g_i$, respectively.
Consequently, the Lagrangian $\mathcal{L}$ is defined as
\begin{equation}\label{eq:lagrangian}
    \mathcal{L}\left(\mathbf{z}, \mathbf{p}\right) 
    = q\left(\mathbf{w}, \mathbf{p}\right) + {\boldsymbol{\nu}}^{\top}\mathbf{h}\left(\mathbf{w}, \mathbf{p}\right) + {\boldsymbol{\lambda}}^{\top} \mathbf{g}\left(\mathbf{w}, \mathbf{p}\right).
\end{equation}

\LL{Add assumptions on regularity (LICQ) and smoothness?}
For $\mathbf{z}^{*}(\mathbf{p})$ to be an optimal solution, the first-order necessary conditions of optimality, known as the Karush-Kuhn-Tucker (KKT) conditions, must be satisfied.
These conditions are given as follows \cite{nocedalNumericalOptimization2006}:
\begin{subequations}\label{eq:kkt_conditions}
    \begin{align}
        \nabla_\mathbf{w} \mathcal{L} \left(\mathbf{w}^{*}, \boldsymbol{\nu}^{*}, \boldsymbol{\lambda}^{*}, \mathbf{p}\right) &= 0, \label{subeq:stationarity_condition} \\
        \mathbf{h}\left(\mathbf{w}^{*}, \mathbf{p}\right) &= 0, \label{subeq:equality_constraint} \\
        \mathbf{g}\left(\mathbf{w}^{*}, \mathbf{p}\right) &\leq 0, \label{subeq:inequality_constraint} \\
        {\boldsymbol{\lambda}^{*}} &\geq 0, \label{subeq:dual_feasibility} \\
        {\boldsymbol{\lambda}^{*}} \odot \mathbf{g}\left(\mathbf{w}^{*}, \mathbf{p}\right) &= 0, \label{subeq:complementary_slackness}
    \end{align}
\end{subequations}
where \eqref{subeq:stationarity_condition} is the stationarity condition, \eqref{subeq:equality_constraint}, \eqref{subeq:inequality_constraint}, and \eqref{subeq:dual_feasibility} represent primal and dual feasibility, and \eqref{subeq:complementary_slackness} is the complementary slackness condition, applying element-wise multiplication (denoted by $\odot$) between the dual variables $\boldsymbol{\lambda}^{*}$ and the inequality constraints $\mathbf{g}\left(\mathbf{w}^{*}, \mathbf{p}\right)$.
If the optimization problem \eqref{eq:nlp} is convex, i.e., the objective function $q$ and the inequality constraints $\mathbf{g}$ are convex, and the equality constraints $\mathbf{h}$ are affine in $\mathbf{w}$ for all $\mathbf{p}$, then these conditions are both necessary and sufficient for optimality, meaning any point satisfying them is a global minimizer \cite{boydConvexOptimization2004,nocedalNumericalOptimization2006}.
For nonconvex problems, KKT points may correspond to local minima, local maxima, or saddle points, and multiple such points may exist. \LL{add second-order sufficient conditions?}

In general, a closed-form solution to problem \eqref{eq:nlp} does not exist.
Therefore, numerical solvers are applied to solve this problem iteratively, typically using the following update rule,
\begin{equation}\label{eq:iterative_solver}
    \mathbf{z}_{k+1} = \mathbf{z}_k + \alpha_k \Delta \mathbf{z}_k,
\end{equation}
where $\mathbf{z}_{k}$ denotes the primal-dual variable at iteration $k$, $\alpha_k$ is the step size, and $\Delta\mathbf{z}_{k}$ is the computed step direction \cite{nocedalNumericalOptimization2006}.
This update is repeated until a stopping criterion, such as a maximum number of iterations or an optimality tolerance, is met.
However, computing these steps can be computationally expensive, especially for large-scale optimization problems.

\subsection{Optimality Condition Reformulation}

The steps $\Delta\mathbf{z}_{k}$ are typically determined by reformulating the KKT conditions \eqref{eq:kkt_conditions} into a nonlinear system of equations $\mathbf{r}(\mathbf{z},\mathbf{p})$, which has its roots at the primal-dual solutions $\mathbf{z}^{*}(\mathbf{p})$, such that $\mathbf{r}(\mathbf{z}^{*}(\mathbf{p});\mathbf{p}) = 0$ and applying Newton or quasi-Newton methods to solve this system \cite{nocedalNumericalOptimization2006}.
Prominent approaches include active-set methods, which iteratively estimate the active constraints and reduce the NLP to an equality constrained problem, and interior-point methods, which relax the non-smooth complementary slackness condition \eqref{subeq:complementary_slackness} and safeguard iterates to satisfy the inequality constraints and dual feasibility \cite{wachterImplementationInteriorpointFilter2006}.
Especially suited for the application to learning-based optimization is the Fischer-Burmeister reformulation, as it does not require updating an active-set and is able to handle infeasible starting points \cite{fischerSpecialNewtontypeOptimization1992,liao-mcphersonFBstabProximallyStabilized2020,lukenLearningIterativeSolvers2024}.
Furthermore, it provides a differentiable system of equations, whose residuals form the basis for the self-supervised training loss function proposed in this work.

A penalized version of the Fischer-Burmeister equation, with better properties with regard to the scale of its derivatives, is defined as the following implicit function that replaces the inequality constraints \eqref{subeq:inequality_constraint}, dual feasibility \eqref{subeq:dual_feasibility} and the complementary slackness \eqref{subeq:complementary_slackness} and implicitly satisfies these equations as long as the residual is zero \cite{chenPenalizedFischerBurmeisterNCPfunction2000,liao-mcphersonFBstabProximallyStabilized2020}:

\begin{align}\label{eq:background_fb_for_nlp}
    \phi_i(\lambda_i, g_i) =\;& \rho\left(\lambda_i - g_i - \sqrt{\lambda_i^2 + g_i^2 + \epsilon^2} \right) \notag \\
    &+ (1-\rho)\lambda_i^+ g_i^- = 0.
\end{align}

Here, $\epsilon$ represents a small smoothing parameter, e.g. $10^{-6}$, and $\rho \in (0,1)$ is a fixed penalization parameter.
Furthermore $\lambda_i^+=\max\{\lambda_i, 0\}$ and $g_i^-=\max\{-g_i, 0\}$ denote the positive parts of $\lambda_i$ and $-g_i$, respectively. 
Every solution of the penalized Fischer-Burmeister equation \eqref{eq:background_fb_for_nlp}  with smoothing parameter $\epsilon = 0$ also satisfies $g_i \leq 0$ and $\lambda_i \geq 0$, as well as the complementary slackness condition $\lambda_i g_i = 0$.
Note that the penalized Fischer-Burmeister function \eqref{eq:background_fb_for_nlp} is semi-smooth, due to the max projections $\lambda_i^+$ and $g_i^-$.
However, this does not affect the subsequent analysis as all assumptions of Lemma \ref{lemma:loss_function_per_sample} continue to hold and standard machine learning libraries such as PyTorch \cite{paszkePyTorchImperativeStyle2019} select a valid generalized Jacobian element on the kink set during backpropagation, analogously to ReLU activation functions. %

For the proof of Lemma \ref{lemma:loss_function_per_sample}, which considers the self-supervised training loss function, we require the derivatives of the penalized Fischer-Burmeister equation \eqref{eq:background_fb_for_nlp} with respect to $\lambda_i$ and $g_i$. \LL{Check theorem reference. maybe refer to lemma and theorem.}
The derivatives of the penalized Fischer-Burmeister equation \eqref{eq:background_fb_for_nlp} with respect to $\lambda_i$ and $g_i$ are given by the following expressions:
\begin{align}\label{eq:background_fb_derivatives_lambda_i}
    \frac{\partial \phi_i(\lambda_i, g_i)}{\partial \lambda_i} 
    &= \rho\left(1 - \frac{\lambda_i}{\sqrt{\lambda_i^2 + g_i^2 + \epsilon^2}} \right) \notag \\
    &\quad + (1-\rho)\mathbb{I}_{\{\lambda_i>0\}} g_i^- , 
\end{align}

\begin{align}\label{eq:background_fb_derivatives_g_i}
    \frac{\partial \phi_i(\lambda_i, g_i)}{\partial g_i} 
    &= \rho\left(-1 - \frac{g_i}{\sqrt{\lambda_i^2 + g_i^2 + \epsilon^2}} \right) \notag \\
    &\quad - (1-\rho)\mathbb{I}_{\{g_i<0\}} \lambda_i^+ . 
\end{align}

Here, $\mathbb{I}_{\{\lambda_i>0\}}$ and $\mathbb{I}_{\{g_i<0\}}$ denote the element-wise indicator functions, which are defined as follows:
\begin{equation}\notag
    \mathbb{I}_{\{\lambda_i>0\}} = \begin{cases}
        1, & \text{if } \lambda_i > 0,\\
        0, & \text{otherwise},
    \end{cases}
    \quad 
    \mathbb{I}_{\{g_i<0\}} = \begin{cases}
        1, & \text{if } g_i < 0,\\
        0, & \text{otherwise}.
    \end{cases}
\end{equation}

For $\epsilon >0$ and $\rho \in (0,1)$, it holds that $\frac{\partial \phi_i(\lambda_i, g_i)}{\partial \lambda_i}>0$ and $\frac{\partial \phi_i(\lambda_i, g_i)}{\partial g_i}<0$ for all $\lambda_i, g_i \in \mathbb{R}$.
\LL{Very Important: This only holds for bounded $\lambda_i$ and $g_i$. For infeasible problems, these can become unbounded.}
Furthermore, these derivatives are continuous and bounded above and below, respectively \cite{liao-mcphersonFBstabProximallyStabilized2020}.

The derivative of $\phi_i(\lambda_i, g_i)$ with respect to the decision variables $\mathbf{w}$ is given by the chain rule as follows:
\begin{equation}
      \frac{\partial \phi_i(\lambda_i, g_i)}{\partial \mathbf{w}} = \frac{\partial \phi_i(\lambda_i, g_i)}{\partial g_i} \frac{\partial g_i}{\partial \mathbf{w}}.
\end{equation}

The nonlinear system of equations representing the optimality conditions then reads as follows:
\begin{equation}\label{eq:r_phi}
    \begin{aligned}
        \mathbf{r}_{\phi}(\mathbf{z}^{*},\mathbf{p})&= \begin{cases}
            \nabla_\mathbf{w} \mathcal{L} \left(\mathbf{w}^{*}, \boldsymbol{\nu}^{*}, \boldsymbol{\lambda}^{*}, \mathbf{p}\right) &= 0, \\
            \mathbf{h}\left(\mathbf{w}^{*}, \mathbf{p}\right) &= 0, \\
            \phi(\boldsymbol{\lambda}^*, \mathbf{g}(\mathbf{w}^*,\mathbf{p})) &= 0,
        \end{cases}
    \end{aligned}    
\end{equation}
where $\phi(\boldsymbol{\lambda}^*, \mathbf{g}(\mathbf{w}^*,\mathbf{p}))$ is the vector of penalized Fischer-Burmeister equations \eqref{eq:background_fb_for_nlp} for all inequality constraints $i = 1, \ldots, n_g$.
The solution of this system of equations is equivalent to the solution of the original KKT conditions \eqref{eq:kkt_conditions} in the limit $\epsilon \rightarrow 0$ \cite{fischerSpecialNewtontypeOptimization1992,liao-mcphersonFBstabProximallyStabilized2020}.

\subsection{Nonlinear Model Predictive Control}

A particularly important class of parametric optimization problems arises in model predictive control, where the parameters $\mathbf{p}$ represent the current system state $\mathbf{x}_{\mathrm{init}}$ and the optimization must be solved repeatedly in real-time.

We consider the following nonlinear discrete-time dynamic system at time step $k$ with states $\mathbf{x}_k \in \mathbb{R}^{n_x}$, control actions $\mathbf{u}_k \in \mathbb{R}^{n_u}$, and dynamics $f: \mathbb{R}^{n_x} \times \mathbb{R}^{n_u} \rightarrow \mathbb{R}^{n_x}$, as well as initial conditions $\mathbf{x}_{\mathrm{init}}$:
\begin{equation}
    \mathbf{x}_{k+1} = f(\mathbf{x}_k,\mathbf{u}_k), \quad \mathbf{x}_0 = \mathbf{x}_{\mathrm{init}}.
\end{equation}

A general nonlinear model predictive control (NMPC) problem for this system is defined as follows \cite{rawlingsModelPredictiveControl2017}:
\begin{subequations}\label{eq:nmpc_problem}
    \begin{align}
        \min_{\mathbf{x}_{[0:N]}, \mathbf{u}_{[0:N-1]}} \quad &V_f \left(\mathbf{x}_N\right) + \sum_{k=0}^{N-1} \ell\left(\mathbf{x}_k, \mathbf{u}_k\right)\\
        \mathrm{s.t.} \quad & \mathbf{x}_{k+1} = f\left(\mathbf{x}_k, \mathbf{u}_k\right), \quad \mathbf{x}_0 = \mathbf{x}_{\mathrm{init}}, \\
        & \mathbf{g}_{\textrm{MPC}}\left(\mathbf{x}_k, \mathbf{u}_k\right) \leq 0,\\
        & \mathbf{h}_{\textrm{MPC}}\left(\mathbf{x}_k, \mathbf{u}_k\right) = 0,\\
        & k \in \{0, \ldots, N-1~|~N \in \mathbb{N}, N \geq 1 \}.
    \end{align}
\end{subequations}

The control horizon is defined by $N \in \mathbb{N}$, where $N \geq 1$, and the decision variables are the state and control trajectories $\mathbf{x}_{[0:N]} = \left(\mathbf{x}_0, \ldots, \mathbf{x}_N\right)$ and $\mathbf{u}_{[0:N-1]} = \left(\mathbf{u}_0, \ldots, \mathbf{u}_{N-1}\right)$.
The cost function consists of a terminal cost $V_f$ at the final time step $N$ and a sum of stage costs $\ell$ over all time steps $k = 0, \ldots, N-1$.
The system dynamics are stated as an equality constraint $\mathbf{x}_{k+1} = f(\mathbf{x}_k, \mathbf{u}_k)$, with the initial state $\mathbf{x}_0 = \mathbf{x}_{\mathrm{init}}$.
Furthermore, the problem includes nonlinear equality and inequality constraints $\mathbf{h}_{\textrm{MPC}}(\mathbf{x}_k, \mathbf{u}_k) = 0$ and $\mathbf{g}_{\textrm{MPC}}(\mathbf{x}_k, \mathbf{u}_k) \leq 0$, respectively, which e.g. can represent state and control bounds.
At each sampling instant, the NMPC problem is solved with the current initial state $\mathbf{x}_{\mathrm{init}}$, and the first control action $\mathbf{u}_0$ is applied to the system.
The NMPC control law can be summarized as follows:
\begin{equation}\label{eq:nmpc_control_law}
    \mathbf{u}_0 = \Pi_{\mathrm{MPC}}(\mathbf{x}_{\mathrm{init}}).
\end{equation}

As solving this optimization problem online can be computationally expensive, often an approximation of the NMPC control law is used, which is based on a neural network approximation of the optimal control action:
\begin{equation}\label{eq:nmpc_approx_control_law}
    \hat{\mathbf{u}}_0 = \Pi_{\mathrm{approxMPC}}(\mathbf{x}_{\mathrm{init}};\theta).    
\end{equation}
This approximate NMPC is often orders of magnitude faster to evaluate than solving the NMPC optimization problem online, as it only requires a single function evaluation of the neural network $\Pi_{\mathrm{approxMPC}}$ with parameters $\theta$.
The parameters $\theta$ of the neural network are typically determined via imitation learning based on a dataset of optimal control actions $\mathbf{u}_0^*$ obtained by solving the NMPC optimization problem \eqref{eq:nmpc_problem} for various initial states $\mathbf{x}_{\mathrm{init}}$ and using standard neural network training techniques \cite{kargEfficientRepresentationApproximation2020,hertneckLearningApproximateModel2018}.

\section{Proposed Method}
\label{sec:method}

\begin{figure*}[t]
    \centering
    \includegraphics[width=.9\linewidth]{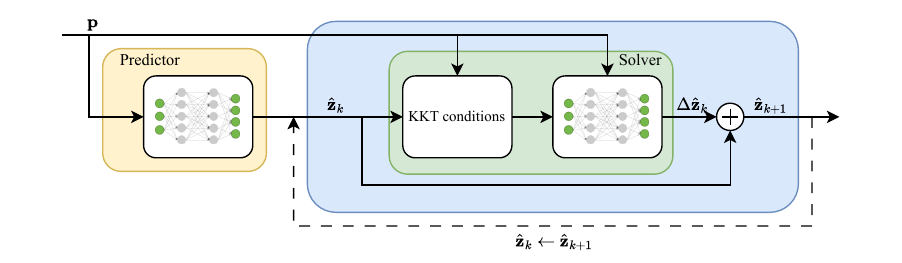}
    \caption{Overview of the LISCO architecture, consisting of a predictor network that provides an initial primal-dual estimate $\hat{\mathbf{z}}_0$ based on the problem parameters $\mathbf{p}$ and a solver network that iteratively refines this estimate by predicting update steps $\Delta \hat{\mathbf{z}}_k$ based on the current primal-dual iterate $\hat{\mathbf{z}}_k$ and the problem parameters $\mathbf{p}$. Both networks are trained in a self-supervised manner using a novel loss function based on the KKT conditions.}
    \label{fig:solver_figure}
\end{figure*}

\subsection{Learning-Based Iterative Solver for Constrained Optimization (LISCO)}
\label{subsec:lisco}
The learning-based iterative solver for constrained optimization (LISCO) is a two-stage approach for solving parametric NLPs and extends our previous work \cite{lukenLearningIterativeSolvers2024}.
It consists of a predictor network that provides an initial estimate of the primal-dual solution from the problem parameters, and a solver network that iteratively refines this estimate until an optimal solution of the desired tolerance is achieved.
The approach uses a novel training loss function based on the KKT conditions for both the predictor and the solver, enabling self-supervised training without prior sampling of optimal solutions for the predictor and solver neural networks. 
The method is designed to be applicable to various types of parametric NLPs, and, by means of a convexification strategy, also to nonconvex problems.
The architecture of LISCO, containing the predictor and solver networks, is illustrated in Fig.~\ref{fig:solver_figure}.

\subsubsection{Predictor}
The predictor consists of a simple feed-forward neural network that maps the problem parameters $\mathbf{p}$ to an initial primal-dual estimate $\hat{\mathbf{z}}_0 = (\hat{\mathbf{w}}_0,\hat{\boldsymbol{\nu}}_0,\hat{\boldsymbol{\lambda}}_0)$.
This allows the solver to start from a good initial guess, which improves convergence speed and training, as shown in Section \ref{sec:case_study}.
The predictor with weights and biases $\theta$ is described below:
\begin{equation}\label{eq:predictor}
    \hat{\mathbf{z}}_0 = \Pi_{\theta,\text{pred}}(\mathbf{p}).
\end{equation}

\subsubsection{Solver}
Similar to approximate MPC \cite{kargEfficientRepresentationApproximation2020} or comparable approaches such as DC3 \cite{dontiDC3LearningMethod2021} or PDL \cite{parkSelfSupervisedPrimalDualLearning2023}, the predictor network only provides an approximation of the optimal solution to \eqref{eq:nlp}.
Therefore, we use this predictor as the starting point for a learned iterative solver that refines the primal-dual estimate $\hat{\mathbf{z}}_0$ until an optimal solution of the desired tolerance is achieved.
The solver network takes the initial primal-dual estimate $\hat{\mathbf{z}}_0$ and refines it iteratively by predicting steps $\Delta \hat{\mathbf{z}}_k  = (\Delta\hat{\mathbf{w}}_k,\Delta\hat{\boldsymbol{\nu}}_k,\Delta\hat{\boldsymbol{\lambda}}_k)$, with which the primal-dual iterates $\mathbf{z}_k$ are updated according to the update rule $\mathbf{z}_{k+1} = \mathbf{z}_k + \alpha_k \Delta \hat{\mathbf{z}}_k$.
Instead of directly predicting the steps based on the primal-dual iterates $\mathbf{z}_k$ and problem parameters $\mathbf{p}$, we utilize a representation based on the KKT conditions \eqref{eq:kkt_conditions}, which is described in detail in Section \ref{sec:implementation_details}.
This representation captures the optimality of the current primal-dual iterate and allows the neural network to focus on correcting the specific violations of the KKT conditions.
The solver network with weights and biases $\theta$ is defined as follows:
\begin{equation}\label{eq:solver}
    \Delta\hat{\mathbf{z}}_{k} = \Pi_{\theta,\text{solv}}(\mathbf{z}_k,\mathbf{p}).
\end{equation}

\subsection{A Novel Loss Function for Self-Supervised Training}
\label{sec:loss_function}
In the following, we present a novel training loss function based on the modified KKT conditions \eqref{eq:r_phi}, which utilize the penalized Fischer-Burmeister equation \eqref{eq:background_fb_for_nlp}.
This loss function is an improvement on the loss function in our previous work \cite{lukenLearningIterativeSolvers2024} and allows both the predictor and the solver to be trained in a self-supervised approach.
This means that no previous optimizer solutions are required for training.
In addition, this loss function is fully GPU-parallelizable, which enables efficient training on modern hardware.
Furthermore, we show in Lemma \ref{lemma:loss_function_per_sample} that, under reasonable assumptions, the loss function only has minima at the KKT points of the optimization problems.
In conjunction with a convexification strategy as described in Section \ref{sec:implementation_details}, this loss function can also be used for the self-supervised training of LISCO on nonconvex optimization problems.

The per-sample loss function is defined as follows:
\begin{equation}\label{eq:loss_function_per_sample}
    l(\mathbf{z}, \mathbf{p}) = \log \left( \frac{1}{2} \lVert \mathbf{r}_{\phi}(\mathbf{z}, \mathbf{p}) \rVert_2^2 + \sigma \right).
\end{equation}

This per-sample loss function is evaluated at a sample of problem parameters $\mathbf{p}^i$ and a corresponding sample of primal-dual predictions $\hat{\mathbf{z}}^{i}$.
The primal-dual predictions $\hat{\mathbf{z}}^{i}$ are dependent on the neural network parameters $\theta$ and are computed differently for predictor and solver networks, as follows:
\begin{equation}\label{eq:z_hat_theta}
    \hat{\mathbf{z}}^{i}(\theta) = \begin{cases}
        \Pi_{\theta,\text{pred}}(\mathbf{p}^i) & \text{for predictor network}, \\
        \mathbf{z}_k^i + \Pi_{\theta,\text{solv}}(\mathbf{z}_k^i, \mathbf{p}^i) & \text{for solver network}.
    \end{cases}
\end{equation}
For the predictor training, the samples $\hat{\mathbf{z}}^{i}$ correspond to the predictions of the predictor network \eqref{eq:predictor} for the given problem parameters $\mathbf{p}^i$ for each sample $i$.
In case of solver training, for given samples of parameters $\mathbf{p}^i$ and primal-dual iterates $\mathbf{z}_k^i$ the samples $\hat{\mathbf{z}}^{i}$ correspond to the updated primal-dual estimates $\hat{\mathbf{z}}_{k+1}^i = \mathbf{z}_k^i + \alpha_k \Delta \hat{\mathbf{z}}_k^{i}$ after applying the predicted steps $\Delta \hat{\mathbf{z}}_k^{i}$ from the solver network \eqref{eq:solver} as defined in update rule \eqref{eq:iterative_solver}.
We use a fixed step size $\alpha_k = 1$ to allow the solver network to learn an appropriate step size implicitly.

The core of the training loss function is the squared 2-norm of the modified KKT residual $\mathbf{r}_{\phi}$ \eqref{eq:r_phi}.
The squared 2-norm of the KKT residual is scaled logarithmically to make the per-sample loss function applicable across several orders of magnitude of KKT residuals.
In addition, a very small offset $\sigma > 0$, e.g. $\sigma = \textrm{1e-16}$, is added in \eqref{eq:loss_function_per_sample} to prevent numerical issues.

During training of the predictor and solver networks, we use mini-batches of $N_{\text{batch}}$ samples of problem parameters $\{\mathbf{p}^i\}_{i=1}^{N_{\text{batch}}}$ and, in case of the solver, primal-dual iterates $\{\mathbf{z}_k^i\}_{i=1}^{N_{\text{batch}}}$, with which the primal-dual predictions $\{\hat{\mathbf{z}}^{i}\}_{i=1}^{N_{\text{batch}}}$ are computed according to \eqref{eq:z_hat_theta}.
The full training loss function is defined as the mean of the per-sample loss functions \eqref{eq:loss_function_per_sample} over all samples in the batch as follows:
\begin{equation}\label{eq:loss_function_full}
    L(\theta) = \frac{1}{N_{\text{batch}}} \sum_{i=1}^{N_{\text{batch}}} l(\hat{\mathbf{z}}^i(\theta), \mathbf{p}^i).
\end{equation}
This loss function can be minimized with respect to the neural network parameters $\theta$ using standard gradient-based optimization methods, such as Adam \cite{kingmaAdamMethodStochastic2017}.
This training approach is completely self-supervised, since no prior optimization solutions are needed to train the predictor or the solver, since the problem parameters $\mathbf{p}^i$ can be sampled randomly from the parameter space and the primal-dual iterates $\mathbf{z}_k^i$ can be generated during training by applying the solver network iteratively starting from random initial guesses.
The effectiveness of this self-supervised training procedure relies on the loss function not having any stationary points other than the KKT points. This property is established in Lemma \ref{lemma:loss_function_per_sample} and Theorem \ref{thm:loss_function_full} of Section \ref{sec:theoretical_properties}, and is a prerequisite for the training procedure to converge to certifiably optimal predictions. %
Further details on the implementation of the training procedure are provided in Section \ref{sec:implementation_details}.

\section{Theoretical Properties of the Loss Function}
\label{sec:theoretical_properties}
A critical prerequisite for self-supervised training of both the predictor and the solver is that the loss function \eqref{eq:loss_function_full} has its minima exclusively at the optimal solutions of the optimization problem.
Specifically, this means that the per-sample loss function \eqref{eq:loss_function_per_sample} is only minimized if the primal-dual estimates $\hat{\mathbf{z}}^i$ satisfy the modified KKT conditions \eqref{eq:r_phi}, i.e., $\mathbf{r}_\phi(\hat{\mathbf{z}}^i, \mathbf{p}^i) = 0$ for all feasible samples $i$.
This is formalized in Lemma \ref{lemma:loss_function_per_sample}.
To this end, we first introduce the following assumptions.

\begin{assumption}\label{assumption:1}
    The parametric optimization problem \eqref{eq:nlp} is feasible for the given parameter instances $\mathbf{p}^i$.
\end{assumption}

\begin{assumption}\label{assumption:2}
    The objective function $q$ and the constraint functions $\mathbf{g}$ and $\mathbf{h}$ are twice continuously differentiable with respect to the decision variables $\mathbf{w}$ for all parameter instances $\mathbf{p}^i$.
\end{assumption}

\begin{assumption}\label{assumption:3}
    The smoothing parameter $\epsilon > 0$ and the offset parameter $\sigma > 0$ are chosen arbitrarily small, but not zero. 
\end{assumption}
\begin{assumption}\label{assumption:4}
    The gradients of the \emph{equality} constraints $\mathbf{h}$ are linearly independent for all samples $(\hat{\mathbf{z}}^{i},\mathbf{p}^i$), i.e., $\nabla_{\mathbf{w}} \mathbf{h}(\mathbf{w}, \mathbf{p})$ has full row rank everywhere.
\end{assumption}
\del{Note that Assumption \ref{assumption:4} only concerns equality constraints and is often satisfied in practice, e.g., in NMPC problems, where the equality constraints represent the system dynamics.}

\begin{assumption}\label{assumption:5}
    The Hessian of the Lagrangian \eqref{eq:lagrangian} with respect to the decision variables $\mathbf{w}$ or its approximation is positive definite for all samples $(\hat{\mathbf{z}}^{i},\mathbf{p}^i$).
\end{assumption}
\del{Assumption \ref{assumption:5} can always be satisfied by choosing a proper Hessian approximation as proposed in the convexification procedure, described in Section \ref{sec:implementation_details}.}

\rev{
Assumptions~\ref{assumption:1}, \ref{assumption:4}, and~\ref{assumption:5} are not restrictive in practice for the targeted applications.
If a parameter instance is infeasible (Assumption~\ref{assumption:1}), the problem admits no optimal solution and the KKT residual is non-zero, so training a solver on such instances is not useful and the per-sample Lemma~\ref{lemma:loss_function_per_sample} does not apply. Practical strategies to handle this are discarding infeasible samples during training by removing stalling samples, as done in this work, sampling parameters from a known feasible domain, e.g.,\ by physical or design considerations, or using a slack-variable reformulation.
Assumption~\ref{assumption:4} requires the equality constraint gradients to be linearly independent.
In typical NMPC formulations with multiple-shooting or direct-collocation discretizations, the state and control variables are independent decision variables with the dynamics and initial conditions being equality constraints.
Therefore, the Jacobian of the equality constraints is structurally of full row rank.
For arbitrary nonlinear equality constraints, degenerate cases must be handled by constraint reformulation or regularization.
Finally, Assumption~\ref{assumption:5} is fulfilled by construction through the positive-definite Hessian approximation defined by the convexification procedure in Section~\ref{sec:convexification}, which is applied only during training.
The approximation shapes the curvature observed during backpropagation, while the deployed iteration in~\eqref{eq:solver_step} evaluates the residuals of the original problem.}

\begin{lemma}\label{lemma:loss_function_per_sample}
Given Assumptions \ref{assumption:1} to \ref{assumption:5}, there exist no other local optima of the per-sample loss function \eqref{eq:loss_function_per_sample} than those points $(\hat{\mathbf{z}}^{i}, \mathbf{p}^{i})$ that satisfy the modified KKT conditions \eqref{eq:r_phi}.
This means that the derivative of the loss function with respect to $\mathbf{z}$ is zero if and only if the modified KKT conditions are satisfied:
\begin{equation}
    \nabla_{\mathbf{z}} l(\hat{\mathbf{z}}^i, \mathbf{p}^i) = 0 \Leftrightarrow \mathbf{r}_\phi(\hat{\mathbf{z}}^i, \mathbf{p}^i) = 0.
\end{equation}
\end{lemma}

\begin{proof}\label{proof:lemma:loss_function_per_sample}
The gradient of the per-sample loss function \eqref{eq:loss_function_per_sample} with respect to the primal-dual variable $\mathbf{z}$ for a sample $(\hat{\mathbf{z}}^i, \mathbf{p}^i)$ is given by:
\begin{equation}\label{eq:sample_grad_loss_function}
    \nabla_{\mathbf{z}} l(\hat{\mathbf{z}}^i, \mathbf{p}^i) 
    = \frac{
        \nabla_{\mathbf{z}} \mathbf{r}_\phi(\hat{\mathbf{z}}^i, \mathbf{p}^i)^\top 
        \mathbf{r}_\phi(\hat{\mathbf{z}}^i, \mathbf{p}^i)
    }{
        \lVert \mathbf{r}_\phi(\hat{\mathbf{z}}^i, \mathbf{p}^i) \rVert_2^2 + \sigma
    }.
\end{equation}

Since $\sigma > 0$ because of Assumption \ref{assumption:3} and $\lVert \mathbf{r}_\phi(\hat{\mathbf{z}}^i, \mathbf{p}^i) \rVert_2^2 \geq 0$ hold, the derivative of the per-sample loss function is equal to zero if and only if
\begin{equation}\label{eq:jvp_kkt_residual}
    \nabla_{\mathbf{z}} l(\hat{\mathbf{z}}^i, \mathbf{p}^i) = 0 \Leftrightarrow \nabla_{\mathbf{z}} \mathbf{r}_\phi(\hat{\mathbf{z}}^i, \mathbf{p}^i)^\top 
    \mathbf{r}_\phi(\hat{\mathbf{z}}^i, \mathbf{p}^i) = 0.
\end{equation}
This is the case when either the residual $\mathbf{r}_\phi(\hat{\mathbf{z}}^i, \mathbf{p}^i)=0$, i.e., the modified KKT conditions are satisfied, or the KKT matrix $\nabla_{\mathbf{z}} \mathbf{r}_\phi(\hat{\mathbf{z}}^i, \mathbf{p}^i)$ is singular.
Since $\mathbf{r}_\phi(\hat{\mathbf{z}}^i, \mathbf{p}^i)=0$ implies that the modified KKT conditions are satisfied, it remains to be shown that the KKT matrix is non-singular under the given assumptions.
The numerator of \eqref{eq:sample_grad_loss_function} is given by
\begin{align}\label{eq:jvp_mod_kkt_matrix}
    &\nabla_{\mathbf{z}} \mathbf{r}_\phi(\hat{\mathbf{z}}^i, \mathbf{p}^i)^\top 
    \mathbf{r}_\phi(\hat{\mathbf{z}}^i, \mathbf{p}^i) \notag \\
    &=
    \begin{bmatrix}
        \mathbf{H}^{\top} & \mathbf{A}^{\top} & \mathbf{B}^{\top}\mathbf{C}\\
        \mathbf{A} & 0 & 0\\
        \mathbf{B}  & 0 & \mathbf{D}
    \end{bmatrix}
    \begin{bmatrix}
        \nabla_{\mathbf{w}} \mathcal{L}(\hat{\mathbf{z}}^{i}, \mathbf{p}^i)\\
        \mathbf{h}(\hat{\mathbf{w}}^i, \mathbf{p}^i)\\
        \boldsymbol{\phi}(\hat{\boldsymbol{\lambda}}^i, \mathbf{g}(\hat{\mathbf{w}}^i, \mathbf{p}^i))
    \end{bmatrix} = 0.
\end{align}

The matrices $\mathbf{A}, \mathbf{B}$, $\mathbf{C}$ and $\mathbf{D}$ are defined as follows:
\begin{align}
    \mathbf{H} & \succ 0,~ \mathbf{H} \in \mathbb{R}^{n_w \times n_w}, \label{eq:def_H}\\
    \mathbf{A} &= \nabla_{\mathbf{w}} \mathbf{h}\left(\hat{\mathbf{w}}^{i}, \mathbf{p}^{i}\right) \in \mathbb{R}^{n_h \times n_w}, \label{eq:def_A}\\
    \mathbf{B} &= \nabla_{\mathbf{w}} \mathbf{g}\left(\hat{\mathbf{w}}^{i}, \mathbf{p}^{i}\right) \in \mathbb{R}^{n_g \times n_w}, \label{eq:def_B}\\
    \mathbf{C} &= \operatorname{diag}\left(\nabla_{\mathbf{g}} \boldsymbol{\phi}(\hat{\boldsymbol{\lambda}}^{i}, \mathbf{g}\left(\hat{\mathbf{w}}^{i}, \mathbf{p}^{i}\right))\right) \in \mathbb{R}^{n_g \times n_g}, \label{eq:def_C} \\
    \mathbf{D} &= \operatorname{diag}\left(\nabla_{\boldsymbol{\lambda}} \boldsymbol{\phi}(\hat{\boldsymbol{\lambda}}^{i}, \mathbf{g}\left(\hat{\mathbf{w}}^{i}, \mathbf{p}^{i}\right))\right) \in \mathbb{R}^{n_g \times n_g}. \label{eq:def_D}
\end{align}
Here, $\mathbf{H}$ is the Hessian of the Lagrangian with respect to the primal variables $\nabla^2_{\mathbf{w}} \mathcal{L}\left(\hat{\mathbf{z}}^{i},\mathbf{p}^{i}\right)$, which exists because of Assumption \ref{assumption:2} or its positive definite approximation (see Section \ref{sec:implementation_details}).
$\mathbf{H}$ is positive definite because of Assumption \ref{assumption:5}.

Because of Assumption \ref{assumption:3}, the matrix $\mathbf{D}$ is a diagonal matrix with strictly positive entries $\frac{\partial \phi_i(\lambda_i, g_i)}{\partial \lambda_i}>0$ and therefore has full rank, is symmetric and is invertible. %
This allows the last row of \eqref{eq:jvp_mod_kkt_matrix} to be eliminated by eliminating the Fischer-Burmeister terms $\boldsymbol{\phi}(\hat{\boldsymbol{\lambda}}^i, \mathbf{g}(\hat{\mathbf{w}}^i, \mathbf{p}^i))$:
\begin{equation}\label{eq:phi_elimination}
    \boldsymbol{\phi}(\hat{\boldsymbol{\lambda}}^i, \mathbf{g}(\hat{\mathbf{w}}^i, \mathbf{p}^i)) = - \mathbf{D}^{-1} \mathbf{B} \nabla_{\mathbf{w}} \mathcal{L}(\hat{\mathbf{z}}^{i}, \mathbf{p}^i).
\end{equation}
This reduces the system of equations \eqref{eq:jvp_mod_kkt_matrix} to:
\begin{equation}\label{eq:reduced_jvp_mod_kkt_matrix}
    \begin{bmatrix}
        \mathbf{S} & \mathbf{A}^{\top}\\
        \mathbf{A} & \mathbf{0}
    \end{bmatrix}
    \begin{bmatrix}
        \nabla_{\mathbf{w}} \mathcal{L}(\hat{\mathbf{z}}^{i}, \mathbf{p}^i)\\
        \mathbf{h}(\hat{\mathbf{w}}^i, \mathbf{p}^i)
    \end{bmatrix}
    = \mathbf{0},
\end{equation}
where $\mathbf{S} = \mathbf{H}^{\top} - \mathbf{B}^{\top}\mathbf{C}\mathbf{D}^{-1}\mathbf{B}$ holds.
We describe the reduced KKT matrix $\mathbf{K}$ and the reduced KKT residual $\tilde{\mathbf{r}}_\phi(\hat{\mathbf{z}}^i, \mathbf{p}^i)$ as follows:
\begin{equation} \label{eq:reduced_kkt_matrix}
    \mathbf{K} = 
    \begin{bmatrix}
        \mathbf{S} & \mathbf{A}^{\top}\\
        \mathbf{A} & \mathbf{0}
    \end{bmatrix}, \quad
    \tilde{\mathbf{r}}_\phi(\hat{\mathbf{z}}^i, \mathbf{p}^i) =
    \begin{bmatrix}
        \nabla_{\mathbf{w}} \mathcal{L}(\hat{\mathbf{z}}^{i}, \mathbf{p}^i)\\
        \mathbf{h}(\hat{\mathbf{w}}^i, \mathbf{p}^i)
    \end{bmatrix}.
\end{equation}
Since $\tilde{\mathbf{r}}_\phi(\hat{\mathbf{z}}^i, \mathbf{p}^i) = \mathbf{0} \Rightarrow \nabla_{\mathbf{w}} \mathcal{L}(\hat{\mathbf{z}}^{i}, \mathbf{p}^i) = \mathbf{0}$ holds, equation \eqref{eq:phi_elimination} implies $\boldsymbol{\phi}(\hat{\boldsymbol{\lambda}}^i, \mathbf{g}(\hat{\mathbf{w}}^i, \mathbf{p}^i)) = \mathbf{0}$.
Therefore, $\tilde{\mathbf{r}}_\phi(\hat{\mathbf{z}}^i, \mathbf{p}^i) = \mathbf{0} \Rightarrow \mathbf{r}_\phi(\hat{\mathbf{z}}^i, \mathbf{p}^i) = \mathbf{0}$ holds.
Consequently, since $\mathbf{r}_\phi(\hat{\mathbf{z}}^i, \mathbf{p}^i) = \mathbf{0} \Rightarrow \tilde{\mathbf{r}}_\phi(\hat{\mathbf{z}}^i, \mathbf{p}^i) = \mathbf{0}$ is implied per definition, it follows that $\tilde{\mathbf{r}}_\phi(\hat{\mathbf{z}}^i, \mathbf{p}^i) = \mathbf{0} \Leftrightarrow \mathbf{r}_\phi(\hat{\mathbf{z}}^i, \mathbf{p}^i) = \mathbf{0}$ holds.
\LL{This is trivial, is it needed?}

Thus, to ensure that the per-sample loss function \eqref{eq:loss_function_per_sample} only has a minimum where $\mathbf{r}_\phi(\hat{\mathbf{z}}^i, \mathbf{p}^i) = 0$ holds, it must be shown that the product of the reduced KKT matrix $\mathbf{K}$ and the reduced residual $\tilde{\mathbf{r}}_\phi(\hat{\mathbf{z}}^i, \mathbf{p}^i)$ is zero if and only if the reduced residual is zero:
\begin{equation}\label{eq:reduced_jvp_mod_kkt_matrix_zero}
    \mathbf{K} \tilde{\mathbf{r}}_\phi(\hat{\mathbf{z}}^i, \mathbf{p}^i) = \mathbf{0} \Rightarrow \tilde{\mathbf{r}}_\phi(\hat{\mathbf{z}}^i, \mathbf{p}^i) = \mathbf{0}.
\end{equation}
This equation is satisfied if the reduced KKT matrix $\mathbf{K}$ is non-singular.
This is the case if both the upper left block matrix $\mathbf{S}$ and the corresponding Schur complement $\mathbf{K}/\mathbf{S}$ are non-singular. 
The Schur complement is given by :
\begin{equation}\label{eq:schur_complement}
    \mathbf{K}/\mathbf{S} = \mathbf{0} - \mathbf{A}\mathbf{S}^{-1}\mathbf{A}^{\top}.
\end{equation}
We introduce $\mathbf{P} = -\mathbf{C}\mathbf{D}^{-1}$, where $\mathbf{P}$ is a diagonal matrix with only positive entries, as the matrices $\mathbf{C}$ and $\mathbf{D}$ are diagonal matrices with strictly negative and positive entries, respectively.
The upper left block $\mathbf{S}$ can therefore be reformulated as follows:
\begin{align}
    \mathbf{S} &= \mathbf{H}^{\top} - \mathbf{B}^{\top}\mathbf{C}\mathbf{D}^{-1}\mathbf{B} \notag \\
               &= \mathbf{H}^{\top} + \mathbf{B}^{\top}\mathbf{P}\mathbf{B}.
\end{align}
Since $\mathbf{H}^{\top}$ is positive definite by Assumption \ref{assumption:5}, i.e., either via strict convexity of the NLP or by applying the convexification procedure described in Section \ref{sec:implementation_details}, and $\mathbf{B}^{\top}\mathbf{P}\mathbf{B}$ is positive semidefinite, $\mathbf{S}$ is positive definite.
This implies that the matrix $\mathbf{S}$ is non-singular.

It must therefore be shown that the Schur complement $\mathbf{K}/\mathbf{S}$ is non-singular.
The matrix $\mathbf{A}$ has full rank by Assumption \ref{assumption:4}, i.e. the equality constraints are linearly independent.
Additionally, $\mathbf{S}^{-1}$ is positive definite as the inverse of a positive definite matrix.
The matrix $\mathbf{A}\mathbf{S}^{-1}\mathbf{A}^{\top}$ is therefore positive definite.
It follows that the Schur complement $\mathbf{K}/\mathbf{S} = - \mathbf{A}\mathbf{S}^{-1}\mathbf{A}^{\top}$ is negative definite and therefore non-singular.
From the invertibility of $\mathbf{S}$ and the Schur complement $\mathbf{K}/\mathbf{S}$ it follows that the reduced KKT matrix $\mathbf{K}$ is non-singular.
This means that the expression \eqref{eq:jvp_mod_kkt_matrix} is only zero if the reduced KKT residual $\tilde{\mathbf{r}}_\phi(\hat{\mathbf{z}}^i, \mathbf{p}^i)$ is zero and correspondingly the per-sample loss function \eqref{eq:loss_function_per_sample} only has a minimum if the modified KKT conditions are fulfilled, i.e. $\mathbf{r}_\phi(\hat{\mathbf{z}}^i, \mathbf{p}^i) = 0$.
\end{proof}

While Lemma \ref{lemma:loss_function_per_sample} shows that the per-sample loss function \eqref{eq:loss_function_per_sample} only has minima at KKT points, additional properties must be considered to ensure the full training loss function \eqref{eq:loss_function_full} is suitable for training the solver and predictor neural networks.
To this end, the full training loss function must exhibit non-vanishing gradients as long as the KKT conditions are not satisfied for all samples in the batch, i.e. the training can continue until the KKT conditions are satisfied for all feasible samples.
This is formalized in Theorem \ref{thm:loss_function_full}, which builds upon Lemma \ref{lemma:loss_function_per_sample} and introduces an additional mild assumption regarding the neural network architecture.

\begin{assumption}\label{assumption:6}
    A linear output layer with trainable per-output biases $\mathbf{b} \in \mathbb{R}^{n_z}$ is used in the neural networks, with $\mathbf{b} \subseteq \theta$.
\end{assumption}
Assumption \ref{assumption:6} is not restrictive and fulfilled by design.
We use this assumption to provide a simple proof of Theorem \ref{thm:loss_function_full}.

\begin{theorem}\label{thm:loss_function_full}
Let $\mathcal{P} \subset \mathbb{R}^{n_p}$ be the closed set of all problem parameters $\mathbf{p}$ of the parametric optimization problem \eqref{eq:nlp}.
Let $\hat{\mathbf{z}} = \hat{\mathbf{z}}(\mathbf{p}, \theta)$ be the primal-dual predictions of the neural network predictor or solver with weights $\theta$ as in \eqref{eq:z_hat_theta}, and let $L_{\mathcal{B}}(\theta)$ be the training loss function \eqref{eq:loss_function_full} computed over a batch $\mathcal{B} \subset \mathcal{P}$ of samples.
Given Assumptions \ref{assumption:1} to \ref{assumption:6}, there exists a non-empty, finite batch $\mathcal{B} \subset \mathcal{P}$ such that $\nabla_\theta L_{\mathcal{B}}(\theta) \neq 0$ if and only if there exists at least one sample $\mathbf{p} \in \mathcal{P}$ with corresponding prediction $\hat{\mathbf{z}}$ that does not satisfy the modified KKT conditions:
$(\exists \mathcal{B} \neq \emptyset: \nabla_\theta L_{\mathcal{B}}(\theta) \neq 0) \Leftrightarrow (\exists \mathbf{p} \in \mathcal{P}: \mathbf{r}_\phi(\hat{\mathbf{z}}, \mathbf{p}) \neq 0).$
\end{theorem}

\begin{proof}
    For a given batch $\mathcal{B}$ of $N_{\text{batch}}$ samples $\{\mathbf{p}^i\}_{i=1}^{N_{\text{batch}}}$ and the corresponding primal-dual predictions $\{\hat{\mathbf{z}}^i\}_{i=1}^{N_{\text{batch}}} $, the gradient of the training loss function with respect to the neural network weights $\theta$ is given by:
    \begin{align}\label{eq:gradient_training_loss}
        \nabla_\theta L_{\mathcal{B}}(\theta) &= \frac{1}{N_{\text{batch}}} \sum_{i=1}^{N_{\text{batch}}} \nabla_\theta l(\hat{\mathbf{z}}^i, \mathbf{p}^i) \notag \\
        &= \frac{1}{N_{\text{batch}}} \sum_{i=1}^{N_{\text{batch}}} \nabla_{\mathbf{z}} l(\hat{\mathbf{z}}^i, \mathbf{p}^i) \frac{\partial \hat{\mathbf{z}}^i}{\partial \theta}.
    \end{align}
    Because of Assumption \ref{assumption:6}, the output of the neural network with linear output layer for sample $i$ is given by:
    \begin{equation}
        \hat{\mathbf{z}}^i = \mathbf{W}\boldsymbol{\psi}^i + \mathbf{b}.
    \end{equation}
    where $\mathbf{W}$ are the weights of the output layer, $\mathbf{b}$ are the biases of the output layer, and $\boldsymbol{\psi}^i$ is the output of the previous hidden layers for sample $i$.
    The derivative of the neural network output with respect to the biases of the output layer then yields the identity matrix:
    \begin{equation}
        \frac{\partial \hat{\mathbf{z}}^i}{\partial \mathbf{b}} = \mathbf{I}_{n_z},
    \end{equation}
    where $\mathbf{I}_{n_z}$ is the identity matrix of dimension $n_z \times n_z$.
    This ensures that the following holds for the derivative of the neural network output with respect to its parameters $\theta$ for all samples $i$: $\frac{\partial \hat{\mathbf{z}}^i}{\partial \theta} \neq \mathbf{0}$.

    For any sample $\mathbf{p}^i$ where $\mathbf{r}_\phi(\hat{\mathbf{z}}^i, \mathbf{p}^i) \neq 0$, we can construct a singleton batch $\mathcal{B} = \{\mathbf{p}^i\}$ with $N_{\text{batch}} = 1$. For this batch, the gradient becomes:
    \begin{equation}
        \nabla_\theta L_{\mathcal{B}}(\theta) = \nabla_{\mathbf{z}} l(\hat{\mathbf{z}}^i, \mathbf{p}^i) \frac{\partial \hat{\mathbf{z}}^i}{\partial \theta}.
    \end{equation}
    The bias component of $\nabla_{\theta} L_{\mathcal{B}}(\theta)$ equals $\nabla_{\mathbf{z}} l(\hat{\mathbf{z}}^i, \mathbf{p}^i)\,\frac{\partial \hat{\mathbf{z}}^i}{\partial \mathbf{b}} = \nabla_{\mathbf{z}} l(\hat{\mathbf{z}}^i, \mathbf{p}^i)\,\mathbf{I}_{n_z} = \nabla_{\mathbf{z}} l(\hat{\mathbf{z}}^i, \mathbf{p}^i) \neq \mathbf{0}$.
    Hence, $\nabla_{\theta} L_{\mathcal{B}}(\theta) \neq \mathbf{0}$.
    
    Therefore, there exists a non-empty batch $\mathcal{B}$ such that $\nabla_\theta L_{\mathcal{B}}(\theta) \neq \mathbf{0}$ if and only if there exists at least one sample $\mathbf{p} \in \mathcal{P}$ with corresponding prediction $\hat{\mathbf{z}}$ that does not satisfy the modified KKT conditions.
\end{proof}

This theorem extends the implications of Lemma \ref{lemma:loss_function_per_sample} to practical training scenarios.
Specifically, it guarantees that, as long as there are parameter instances where the neural network's predictions do not satisfy the KKT conditions, there will always be training batches with non-zero gradients, allowing the neural network training to continue.
Together, Lemma \ref{lemma:loss_function_per_sample} (in primal-dual space) and Theorem \ref{thm:loss_function_full} (in parameter space) are a prerequisite for self-supervised training. Without them, gradient descent on \eqref{eq:loss_function_full} could, in principle, stall at predictions that do not satisfy the KKT conditions. %
\rev{Importantly, Lemma~\ref{lemma:loss_function_per_sample} and Theorem~\ref{thm:loss_function_full} provide guarantees on the training loss, ensuring that if the training procedure converges, the learned predictions satisfy the KKT conditions.
They do not guarantee that the learned iterative update rule converges from arbitrary initial iterates at inference time.
Furthermore, for nonconvex problems, KKT conditions are necessary but not sufficient for global optimality, and stationary points may correspond to local minima, saddle points, or local maxima.}

\section{Implementation Details}
\label{sec:implementation_details}

\subsection{Solver Network Architecture}

\begin{figure}
    \centering
    \includegraphics[width=.9\columnwidth]{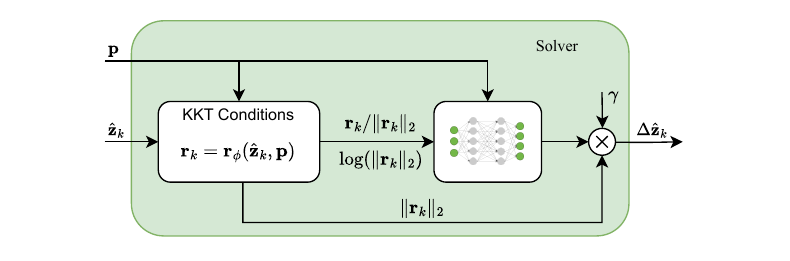}
    \caption{Architecture of the solver network (detailed view from Fig.~\ref{fig:solver_figure}). The neural network $\Psi_{\theta}$ takes normalized KKT residuals $\boldsymbol{\tau}_k$ and problem parameters $\mathbf{p}$ as inputs, where $\boldsymbol{\tau}_k$ contains both the violation direction and magnitude information (equation \eqref{eq:tau}). The network output is scaled by the KKT residual 2-norm $\|\mathbf{r}_k\|_2$ and a problem-specific factor $\gamma$ to produce steps $\Delta \hat{\mathbf{z}}_k$ (equation \eqref{eq:solver_step}).}
    \label{fig:solver_zoom_figure}
\end{figure}

Instead of directly predicting steps $\Delta \hat{\mathbf{z}}_k$ based on the primal-dual iterates $\mathbf{z}_k$ and problem parameters $\mathbf{p}$, both the input and output of the neural network utilize the modified KKT conditions \eqref{eq:r_phi}, which are characterized by the residual $\mathbf{r}_k = \mathbf{r}_\phi(\hat{\mathbf{z}}_k,\mathbf{p})$.
The architecture of the solver network is illustrated in Fig.~\ref{fig:solver_zoom_figure}.

The KKT residuals can span multiple orders of magnitude, making them unsuitable for direct use without preprocessing.
To this end, we construct the input vector $\boldsymbol{\tau}_k$ by normalizing the KKT residual $\mathbf{r}_k$ by its 2-norm to obtain the violation direction, and separately encoding the residual magnitude through its logarithm.
The normalized input is defined as:
\begin{equation}\label{eq:tau}
    \boldsymbol{\tau}_k = \boldsymbol{\tau}(\mathbf{z}_k,\mathbf{p}) = (\mathbf{r}_k/\lVert \mathbf{r}_k \rVert_2, \log(\lVert \mathbf{r}_k \rVert_2)).
\end{equation}

To ensure that the solver network can predict steps with sizes across multiple orders of magnitude, the neural network output is scaled by the 2-norm of the KKT residual $\|\mathbf{r}_k\|_2$.
This enables both small corrections close to the optimal solution $\mathbf{z}^*$ and large steps for distant primal-dual iterates $\mathbf{z}_k$.
Additionally, a constant scaling factor $\gamma > 0$ is introduced for problem-specific scaling.

The neural network that forms the core of the solver architecture is denoted by $\Psi_{\theta}$, where $\theta$ denotes the weights and biases of the neural network.
The steps $\Delta \hat{\mathbf{z}}_k$ are then predicted as follows:
\begin{equation}\label{eq:solver_step}
    \Delta\hat{\mathbf{z}}_{k} = \Pi_{\theta,\text{solv}}(\mathbf{z}_k,\mathbf{p}) = \gamma \lVert \mathbf{r}_k \rVert_2 \Psi_{\theta}(\boldsymbol{\tau}_k,\mathbf{p}).
\end{equation}

The step size $\alpha_k$ \eqref{eq:iterative_solver} plays a different role in LISCO than in classical iterative solvers.
Since the loss function \eqref{eq:loss_function_full} of the neural network is based on the KKT conditions at the predicted primal-dual solution, and thus the size of the steps $\Delta \hat{\mathbf{z}}_k$ is directly trained, a step size of $\alpha_k = 1$ should ideally be used. 
Only when the conditioning of the optimization problem is very poor, it may be useful to reduce the step size $\alpha_k$ within the framework of a line-search in order to improve convergence.

\subsection{Convexification Procedure}
\label{sec:convexification}

The optimization problem \eqref{eq:nlp} is generally nonconvex.
Therefore, KKT points may be local minima, saddle points, or maxima \cite{nocedalNumericalOptimization2006}.
To guide the predictions toward the local minimizers and satisfy the positive-definiteness assumption of Lemma \ref{lemma:loss_function_per_sample}, the loss is convexified using a local quadratic model expanded around the predicted iterate $\hat{\mathbf{z}}$ at each training step.
Because this point coincides with the prediction (i.e., $\bar{\mathbf{w}}=\hat{\mathbf{w}}$ by construction), the convexified residual is equal to the original residual: $\bar{\mathbf{r}}_{\phi}(\hat{\mathbf{z}},\mathbf{p};\bar{\mathbf{w}}) = \mathbf{r}_{\phi}(\hat{\mathbf{z}},\mathbf{p})$.
Thus, no approximation of the NLP is introduced.
The convexification shapes the curvature observed during backpropagation and is discarded during inference, where LISCO uses the residuals of \eqref{eq:nlp}, as in \eqref{eq:solver_step}.
\rev{This convexification can therefore be interpreted as a training-time curvature regularization.
It introduces a deliberate bias on the gradient signal used in backpropagation toward those of a convex local model, but it leaves the residual itself, and therefore the inference behavior, unchanged.}

Formally, the local quadratic model is constructed analogously to sequential quadratic programming (SQP) methods \cite{nocedalNumericalOptimization2006}.
We introduce the variable $\Delta \mathbf{w} := \mathbf{w} - \bar{\mathbf{w}}$ to simplify notation.
The quadratic approximation of the NLP around $\bar{\mathbf{w}}$ then becomes:
\begin{subequations}\label{eq:convexification}
\begin{align}
    \min_{\Delta \mathbf{w}} \quad & q(\bar{\mathbf{w}}, \mathbf{p}) + \nabla_{\mathbf{w}} q(\bar{\mathbf{w}}, \mathbf{p})^\top \Delta \mathbf{w} + \frac{1}{2} \Delta \mathbf{w}^\top \mathbf{H} \Delta \mathbf{w} \\
    \mathrm{s.t.} \quad & \mathbf{g}(\bar{\mathbf{w}}, \mathbf{p}) + \nabla_{\mathbf{w}} \mathbf{g}(\bar{\mathbf{w}}, \mathbf{p})^\top \Delta \mathbf{w} \leq \mathbf{0}, \\
    & \mathbf{h}(\bar{\mathbf{w}}, \mathbf{p}) + \nabla_{\mathbf{w}} \mathbf{h}(\bar{\mathbf{w}}, \mathbf{p})^\top \Delta \mathbf{w} = \mathbf{0}.
\end{align}
\end{subequations}
The objective function $q(\mathbf{w},\mathbf{p})$ of the original optimization problem \eqref{eq:nlp} is approximated locally quadratically, where $\mathbf{H}$ is any positive definite approximation of the Hessian of the Lagrangian with respect to the primal variables $\nabla^2_{\mathbf{w}} \mathcal{L}$ and thus the quadratic approximation is convex.
In the case of a strictly convex optimization problem, $\mathbf{H} = \nabla^2_{\mathbf{w}} \mathcal{L}$ and no further approximation is necessary.
The constraints are linearized.
Due to the affine constraints and the convex objective function, the resulting optimization problem is strictly convex and has exactly one global minimum that satisfies the KKT conditions \cite{nocedalNumericalOptimization2006}.

The modified KKT residuals for the convexified problem \eqref{eq:convexification} at the point $\bar{\mathbf{w}}$ are given by:
\begin{equation}\label{eq:convexified_r_phi}
    \begin{aligned}
        &\bar{\mathbf{r}}_{\phi}(\mathbf{z},\mathbf{p};\bar{\mathbf{w}}) \\
        &=
        \begin{cases}
            \nabla_{\mathbf{w}} \mathcal{L}(\bar{\mathbf{w}},\boldsymbol{\nu},\boldsymbol{\lambda},\mathbf{p}) + \mathbf{H}(\mathbf{w}-\bar{\mathbf{w}}),\\
            \mathbf{h}\left(\bar{\mathbf{w}}, \mathbf{p}\right) + \nabla_{\mathbf{w}} \mathbf{h}\left(\bar{\mathbf{w}}, \mathbf{p}\right)^\top \left(\mathbf{w}-\bar{\mathbf{w}}\right),\\
                        \boldsymbol{\phi}\left(\boldsymbol{\lambda}, \mathbf{g}\left(\bar{\mathbf{w}}, \mathbf{p}\right) + \nabla_{\mathbf{w}} \mathbf{g}\left(\bar{\mathbf{w}}, \mathbf{p}\right)^\top \left(\mathbf{w}-\bar{\mathbf{w}}\right)\right).
        \end{cases}
    \end{aligned}
\end{equation}
Here, $\nabla_{\mathbf{w}} \mathcal{L}(\bar{\mathbf{w}},\boldsymbol{\nu},\boldsymbol{\lambda},\mathbf{p}) = \nabla_{\mathbf{w}} q(\bar{\mathbf{w}},\mathbf{p}) +  \boldsymbol{\nu}^\top\nabla_{\mathbf{w}} \mathbf{h}(\bar{\mathbf{w}},\mathbf{p}) +  \boldsymbol{\lambda}^\top\nabla_{\mathbf{w}}\mathbf{g}(\bar{\mathbf{w}},\mathbf{p})$ describes the gradient of the Lagrangian with respect to the primal variables $\mathbf{w}$ at the point $\bar{\mathbf{w}}$ and the dual variables $\boldsymbol{\nu}$ and $\boldsymbol{\lambda}$. 
For $\mathbf{w} = \bar{\mathbf{w}}$, we therefore have $\bar{\mathbf{r}}_{\phi}(\mathbf{z},\mathbf{p};\bar{\mathbf{w}}) = \mathbf{r}_{\phi}(\mathbf{z},\mathbf{p})$, since $\Delta \mathbf{w} = \mathbf{w} - \bar{\mathbf{w}} = 0$.
During training, convexification is applied at each predicted point $\hat{\mathbf{z}}$, so the equality $\hat{\mathbf{w}} = \bar{\mathbf{w}}$ holds throughout, and the local equivalence stated above follows.

Furthermore, by applying this convexification strategy, the modified KKT matrix of the convexified problem formulation $\nabla_{\mathbf{z}} \bar{\mathbf{r}}_{\phi}(\hat{\mathbf{z}},\mathbf{p};\bar{\mathbf{w}})$ has the same structure as the modified KKT matrix of the original problem formulation $\nabla_{\mathbf{z}} \mathbf{r}_{\phi}(\hat{\mathbf{z}},\mathbf{p})$ in \eqref{eq:jvp_mod_kkt_matrix}.
The only difference is that the Hessian of the Lagrangian with respect to the primal variables $\nabla^2_{\mathbf{w}} \mathcal{L}(\hat{\mathbf{z}},\mathbf{p})$ is replaced by the positive definite matrix $\mathbf{H}$.
\rev{
While this regularization guides the network toward minimizers during training, it does not provide a general guarantee that the solver only converges to local minimizers for arbitrary nonconvex problems.
In nonconvex settings, KKT points may still correspond to local minima, saddle points, or maxima.
Stronger guarantees would require second-order checks or problem-specific safeguards, which are beyond the scope of the presented method.
}

\subsection{Self-Supervised Training}
\label{sec:self_supervised_training}
The self-supervised training of LISCO consists of two steps, the training of the predictor network and the subsequent training of the solver network.

During predictor training, a random batch with $N_\text{batch}$ problem parameters $\{\mathbf{p}^{i}\}_{i=1}^{N_\text{batch}}$ is sampled at each training step, e.g., via uniform random sampling.
The predictor $\Pi_{\theta,\text{pred}}$ \eqref{eq:predictor} is then evaluated so that primal-dual estimates $\{\hat{\mathbf{z}}^{i}\}_{i=1}^{N_\text{batch}}$ are determined.
The previously established loss function \eqref{eq:loss_function_full} is used and the parameters $\theta$ of the predictor network are optimized e.g. via stochastic gradient-based optimization methods, such as Adam \cite{kingmaAdamMethodStochastic2017}.
This training approach is completely self-supervised, since no prior optimization solutions are needed to train the predictor as the loss function is based solely on the sampled problem parameters $\mathbf{p}^i$ and the predictions $\hat{\mathbf{z}}^i$.
In addition, the problem parameters $\mathbf{p}^i$ can be resampled completely from the relevant parameter space in each training step, so that a very large coverage of the parameter space can be achieved, especially in combination with GPU hardware and large batch sizes.
\rev{The choice of parameter sampling distribution directly shapes the regime in which LISCO is trained to perform well.
Uniform random sampling over the known feasible parameter domain, as used in the case studies of Section~\ref{sec:case_study}, is a conservative default that provides broad coverage without requiring prior knowledge of the deployment distribution.
Performance on parameter instances substantially outside the training domain is not expected to be reliable.}

The training of the solver network is similar in principle to the training of the predictor, but has some important special features.
For a given batch of primal-dual estimates $\{\mathbf{z}_k^i\}_{i=1}^{N_{\text{batch}}}$ and problem parameters $\{\mathbf{p}^i\}_{i=1}^{N_{\text{batch}}}$, the solver network \eqref{eq:solver_step} is evaluated to calculate solver steps $\{\Delta \hat{\mathbf{z}}_k^i\}_{i=1}^{N_{\text{batch}}}$.
These are then used via equation \eqref{eq:iterative_solver} with a step size of $\alpha_k = 1$ to determine updated primal-dual estimates $\hat{\mathbf{z}}_{k+1}^i = \mathbf{z}_k^i + \Delta \hat{\mathbf{z}}_k^i$.
The loss function \eqref{eq:loss_function_full} is then evaluated at the updated primal-dual estimates $\hat{\mathbf{z}}_{k+1}^i$ and the parameters $\theta$ of the solver network are optimized accordingly.

To initialize the training, the problem parameters $\mathbf{p}^i$ are randomly sampled from the relevant parameter space. 
The initial primal-dual estimates $\{\hat{\mathbf{z}}_0^i\}_{i=1}^{N_{\text{batch}}}$ can either be sampled randomly or, as recommended in this work, generated from the primal-dual estimates of the pretrained predictor. 
Using the predictor as initialization improves both training and inference speed, as significantly better starting values are available.

A key challenge in training a convergent solver is that the solver network must process primal-dual estimates over a wide range of KKT residual magnitudes during training.
This includes both iterates close to the KKT points and those far away from them in order to ensure robust functionality across the entire range.
However, at the beginning of training, the initial primal-dual iterates only cover a limited part of this range, as they typically have large KKT residuals.

To address this problem, a self-supervised sampling algorithm is used, which dynamically adjusts the training distribution of the primal-dual estimates during training.
The predicted steps $\Delta \hat{\mathbf{z}}_k^i$ of the solver are used to iteratively update the primal-dual estimates.
Once a maximum number of iterations $N_{\text{max}}$ is reached or the KKT residuals $\mathbf{r}_\phi(\hat{\mathbf{z}}_{k+1}^i, \mathbf{p}^i)$ fall below a specified tolerance $\delta$, the corresponding sample $\mathbf{p}^i$ is resampled and the primal-dual estimates are reset accordingly.
This adaptive approach enables the solver network to learn a wide range of KKT residuals and to consider both iterates close to the KKT points and those further away.
During training, the predicted steps continuously improve, so that the solver gradually requires fewer iterations to satisfy the KKT conditions.
The training of the solver network is summarized in Algorithm \ref{alg:solver_training}.
\rev{Together, the logarithmic loss scaling, the offset $\sigma$, the residual-norm output scaling, and the resampling strategy stabilize self-supervised training across the full range of KKT residual magnitudes encountered during training.}

\begin{algorithm}[t]
\caption{Self-Supervised Solver Training}
\label{alg:solver_training}
\begin{algorithmic}[1]
\REQUIRE Pretrained predictor $\Pi_{\theta,\text{pred}}$, batch size $N_{\text{batch}}$, tolerance $\delta$, max iterations $N_{\text{max}}$
\STATE Sample batch of parameters $\{\mathbf{p}^i\}_{i=1}^{N_{\text{batch}}}$
\STATE Initialize $\hat{\mathbf{z}}_0^i \leftarrow \Pi_{\theta,\text{pred}}(\mathbf{p}^i)$ for all $i$ \COMMENT{Eq. \eqref{eq:predictor}}

\FOR{training epochs}
    \STATE \textbf{Compute Solver Steps:}
    \STATE $\Delta\hat{\mathbf{z}}_k^i \leftarrow \Pi_{\theta,\text{solv}}(\hat{\mathbf{z}}_k^i, \mathbf{p}^i)$ for all $i$ \COMMENT{Eq. \eqref{eq:solver_step}}
    \STATE $\hat{\mathbf{z}}_{k+1}^i \leftarrow \hat{\mathbf{z}}_k^i + \Delta\hat{\mathbf{z}}_k^i$ with $\alpha = 1$ \COMMENT{Eq. \eqref{eq:iterative_solver}}
    
    \STATE \textbf{Update Network:}
    \STATE Compute loss: $L(\theta) = \frac{1}{N_{\text{batch}}} \sum_{i=1}^{N_{\text{batch}}} l(\hat{\mathbf{z}}_{k+1}^i, \mathbf{p}^i)$ \COMMENT{Eq. \eqref{eq:loss_function_full}}
    \STATE Update $\theta$ using stochastic gradient descent methods (e.g., Adam)
    
    \STATE \textbf{Update Iterates and Resample:}
    \FOR{each instance $i$}
        \STATE $\hat{\mathbf{z}}_k^i \leftarrow \hat{\mathbf{z}}_{k+1}^i$
        \IF{$\|\bar{\mathbf{r}}_\phi(\hat{\mathbf{z}}_k^i, \mathbf{p}^i)\|_2^2 < \delta$ or $k \geq N_{\text{max}}$}
            \STATE Resample $\mathbf{p}^i$, reset $\hat{\mathbf{z}}_0^i \leftarrow \Pi_{\theta,\text{pred}}(\mathbf{p}^i)$, $k \leftarrow 0$
        \ENDIF
    \ENDFOR
\ENDFOR
\RETURN Trained solver $\Pi_{\theta,\text{solv}}$
\end{algorithmic}
\end{algorithm}

\LL{Add subsection for two-way backtracking line-search?}

\section{Numerical Examples}
\label{sec:case_study}

\subsection{NMPC of a Nonlinear Double Integrator}
To evaluate the effectiveness of LISCO, particularly in comparison to exact optimizer solutions and approximate MPC, we consider the NMPC of a nonlinear double integrator, adapted from \cite{lazarInputtostateStabilityMin2008}.
The system with two states $\mathbf{x} \in \mathbb{R}^2$ and one control action $\mathbf{u} \in \mathbb{R}$ is described by the following discrete-time nonlinear dynamical model:

\begin{align}\label{eq:cs_nldi_model}
    \mathbf{x}_{k+1} &= 
    \begin{bmatrix}
        1 & 1\\
        0 & 1
    \end{bmatrix}
    \mathbf{x}_k   
    +
    \begin{bmatrix}
        0.5\\
        1
    \end{bmatrix}
    \mathbf{u}_k
    +
    \begin{bmatrix}
        0.025\\
        0.025
    \end{bmatrix}
    \mathbf{x}_k^{\intercal}\mathbf{x}_k \notag\\
        &= f(\mathbf{x}_k,\mathbf{u}_k).
\end{align}

The NMPC problem considers the constraints $\lVert \mathbf{x}_k \rVert_{\infty} \leq 10$ and $\lVert \mathbf{u}_k \rVert_{\infty} \leq 2$ on the states and control actions, respectively.
The objective is to minimize the distance to the origin while penalizing the control effort.
To this end, the stage cost $\ell\left(\mathbf{x}_k, \mathbf{u}_k\right)$ and terminal cost $V_f(\mathbf{x}_N)$ are defined as follows:
\begin{equation}
    \ell\left(\mathbf{x}_k, \mathbf{u}_k\right)=
    \mathbf{x}_k^{\intercal}\mathbf{Q}
    \mathbf{x}_k + \mathbf{u}_k^{\intercal}\mathbf{R}\mathbf{u}_k,
\end{equation}
\begin{equation}
    V_f\left(\mathbf{x}_N\right) =
    \mathbf{x}_N^{\intercal}\mathbf{Q}
    \mathbf{x}_N,
\end{equation}
with $\mathbf{Q} = \begin{pmatrix}
		0.8 & 0 \\
		0 & 0.8
	\end{pmatrix} \in \mathbb{R}^{2 \times 2}$ and $\mathbf{R} = 0.1 \in \mathbb{R}$.

The finite-time NMPC problem with horizon $N$ is then formulated as follows:
\begin{subequations}\label{eq:cs_nmpc}
    \begin{align}
        \min_{\mathbf{x}_{[0:N]}, \mathbf{u}_{[0:N-1]}} \quad &V_f \left(\mathbf{x}_N\right) + \sum_{k=0}^{N-1} \ell\left(\mathbf{x}_k, \mathbf{u}_k\right)\\
        \mathrm{s.t.} \quad & \mathbf{x}_{k+1} = f\left(\mathbf{x}_k, \mathbf{u}_k\right),
        \quad \mathbf{x}_0 = \mathbf{x}_{\mathrm{init}}, \\
        & \lVert\mathbf{x}_{k+1} \rVert_{\infty} \leq 10,\\
        & \lVert \mathbf{u}_k \rVert_{\infty} \leq 2, \\
        & k \in \{0, \ldots, N-1~|~N \in \mathbb{N}, N \geq 1 \}.
    \end{align}
\end{subequations}

The parameters of this optimization problem are the current initial states $\mathbf{x}_{\mathrm{init}}$.
Therefore, the parameter vector is defined as $\mathbf{p} = \mathbf{x}_{\mathrm{init}} \in \mathbb{R}^2$, with the dimension $n_p = 2$.
The decision variables of the optimization problem can be summarized as $\mathbf{w} = \left(\mathbf{x}_{[0:N]}, \mathbf{u}_{[0:N-1]}\right) \in \mathbb{R}^{(N+1)\cdot n_x + N\cdot n_u}$, where $n_x = 2$ and $n_u = 1$.
For a prediction horizon of $N=10$, this results in $n_w = (N+1)\cdot n_x + N\cdot n_u = 32$.
The number of Lagrange multipliers for the equality and inequality constraints is $n_{\nu} = (N+1)\cdot n_x = 22$ and $n_{\lambda} = N\cdot2n_x + N\cdot2n_u = 60$.
The total dimension of the primal-dual solution $\mathbf{z} \in \mathbb{R}^{n_z}$ follows as $n_z = n_w + n_{\nu} + n_{\lambda} = 114$.

This optimization problem is nonconvex due to the nonlinear dynamics \eqref{eq:cs_nldi_model} and therefore the convexification strategy described in \eqref{eq:convexification} is applied.
Since the nonconvexity originates from the nonlinear dynamics and the objective function, i.e. the sum of stage and terminal cost, is strictly positive definite, an apparent choice is to substitute the Hessian of the Lagrangian with the Hessian of the objective function, which is defined as follows:
\begin{equation}\label{eq:cs_nldi_hessian}
    \mathbf{H} = \begin{bmatrix}
        \mathbf{Q} & & & & & \\
        & \ddots & & & & \\
        & & \mathbf{Q} & & & \\
        & & & \mathbf{R} & & \\
        & & & & \ddots & \\
        & & & & & \mathbf{R}
    \end{bmatrix} \in \mathbb{R}^{n_w \times n_w},
\end{equation}
where the block-diagonal matrix contains $N+1$ blocks of $\mathbf{Q} \in \mathbb{R}^{2 \times 2}$ corresponding to the state cost matrices and $N$ blocks of $\mathbf{R} \in \mathbb{R}^{1 \times 1}$ corresponding to the control cost matrices.

First, a neural network predictor was trained.
Subsequently, Algorithm~\ref{alg:solver_training} was used to train a neural network solver that leverages the predictor for generating initial iterates.
Additionally, another solver was trained using random initial iterates instead.
In all cases, a smoothing and penalization parameter of $\epsilon = \textrm{1e-8}$ and $\rho = 0.8$ were used in the Fischer-Burmeister function \eqref{eq:background_fb_for_nlp}.
The parameters $\mathbf{p}^{i}=\mathbf{x}_{\mathrm{init}}^{i}$ were sampled uniformly at random within the state constraints.
For solver training, a tolerance of $\delta = \textrm{1e-16}$ for the squared 2-norm in the loss function \eqref{eq:loss_function_per_sample} and a maximum number of iterations of $N_{\text{max}} = 2000$ were applied.
A problem-specific scaling factor of $\gamma = 0.1$ was used in \eqref{eq:solver_step}.
Additionally, individual data points $\mathbf{p}^{i}$ and their corresponding primal-dual estimates $\hat{\mathbf{z}}_k^{i}$ were resampled if no improvement in the KKT residuals $\mathbf{r}_\phi(\hat{\mathbf{z}}_k^i, \mathbf{p}^i)$ was observed during training after $N_{\text{resample}} = 5$ iterations and the 2-norm of the KKT residuals $\lVert \mathbf{r}_k \rVert_2$ used for the scaling of the neural network output in \eqref{eq:solver_step} was clipped to a maximum value of $1.0$ during training.
This strategy prevents iterates from diverging too far, especially during early training, which could lead to numerical instabilities.
Additionally, as some points in the parameter space lead to infeasible NMPC problems, the problem might become ill-conditioned during training.
To address this, we add a simple regularization term during training similar to the approach in \cite{liao-mcphersonFBstabProximallyStabilized2020}.
Specifically, we modify the equality and inequality constraint terms in the KKT residuals \eqref{eq:r_phi} and \eqref{eq:convexified_r_phi} as follows:
\begin{align}
    \mathbf{h}(\mathbf{w}_k, \mathbf{p}) &\to \mathbf{h}(\mathbf{w}_k, \mathbf{p}) - \omega (\boldsymbol{\nu}_k - \bar{\boldsymbol{\nu}}_k), \\
    \mathbf{g}(\mathbf{w}_k, \mathbf{p}) &\to \mathbf{g}(\mathbf{w}_k, \mathbf{p}) - \omega (\boldsymbol{\lambda}_k - \bar{\boldsymbol{\lambda}}_k),
\end{align}
where $\omega = \textrm{1e-2}$ is a regularization parameter.
Here, $\bar{\boldsymbol{\nu}}_k$ and $\bar{\boldsymbol{\lambda}}_k$ are fixed copies of the dual variables: they equal the current dual variables during the forward pass ($\boldsymbol{\lambda}_k = \bar{\boldsymbol{\lambda}}_k$ and $\boldsymbol{\nu}_k = \bar{\boldsymbol{\nu}}_k$), but are treated as constants during backpropagation (i.e., gradients do not flow through them).
This improves the conditioning of the KKT matrix $\nabla_{\mathbf{z}} \mathbf{r}_{\phi}(\hat{\mathbf{z}}_k, \mathbf{p})$ during training, especially when the dual variables become large in magnitude.

For comparison, an approximate MPC was trained using supervised learning.
To this end, a training dataset with $N_{\text{train}} = 25000$ data points was generated in a closed-loop fashion.
Starting from initial states $\mathbf{x}_{\mathrm{init}}$, which were sampled uniformly at random within the state constraints \eqref{eq:cs_nmpc}, the NMPC problem \eqref{eq:nmpc_problem} was solved with the nonlinear optimizer IPOPT \cite{wachterImplementationInteriorpointFilter2006} with an optimizer tolerance of $\textrm{1e-6}$, and the computed control actions $\mathbf{u}_k$ were applied to the system dynamics \eqref{eq:cs_nldi_model} to compute the next states $\mathbf{x}_{k+1}$.
This was repeated for $N_{\mathrm{sim}}=5$ simulation steps to obtain a trajectory, where each data pair $\left(\mathbf{x}_{k}, \mathbf{u}_k\right)$ was included in the training dataset.
The NMPC problem was implemented using CasADi \cite{anderssonCasADiSoftwareFramework2019}.
Trajectories that violate state or input constraints in at least one simulation step were discarded. \LL{Add info an how many data points are discarded.}
For training the neural networks, Adam \cite{kingmaAdamMethodStochastic2017} with the amsgrad variant \cite{reddiConvergenceAdam2019} was used in all cases.
The hyperparameters that were applied for training the predictor, solvers and approximate MPC are summarized in Table~\ref{tab:hyperparameters_nldi}.
The GeLU activation function \cite{hendrycksGaussianErrorLinear2016} was used in all cases for the hidden layers, along with a linear activation in the output layer.
The training was implemented in PyTorch \cite{paszkePyTorchImperativeStyle2019} and performed on an NVIDIA RTX 4090 GPU.

\begin{table}[t]
    \caption{Hyperparameters and training times for the predictor, solvers, and approximate MPC on the NMPC problem.}
    \label{tab:hyperparameters_nldi}
    \begin{threeparttable}
        \centering
        \begin{tabular*}{\tblwidth}{@{} LCCC@{} }
\toprule
 & Approx. & Predictor & Solver\\
 & MPC &  & w. Predictor \\
\midrule
Hidden Layers & 4 & 4 & 4 \\
Neurons per Layer & 512 & 512 & 512 \\
Training Dataset Size & 25000 & - & - \\
Batch Size (per Epoch) & 4096 (7) & 4096 (1) & 4096 (1) \\
Epochs & 50k & 50k & 500k \\
Learning Rate & 1e-03 & 1e-03 & 1e-03 \\
\midrule
Training Time [min] & 9.43 (15.96)\tnote{a} & 1.48 & 21.72 (23.20)\tnote{b} \\
\bottomrule
\end{tabular*}

        \begin{tablenotes}[flushleft]
            \item [a] \del{The training data sampling time is added in parentheses.}\rev{The time needed to generate the training data is added in parentheses.}
            \item [b] The predictor training time is added in parentheses.
        \end{tablenotes}
    \end{threeparttable}
\end{table}

When applying the solver, we also employed a two-way backtracking line-search that reduces the step size $\alpha$ by a factor of $0.95$ when the KKT residuals of the current iteration are at least a factor of $10$ larger than the best KKT residuals achieved so far.
If the KKT residuals are reduced again through the iterations, the step size is reset.
Since the solver is trained such that the predicted steps have a step size of $\alpha=1$, this approach is only necessary in a fraction ($\leq 1\%$) of cases when the problem is ill-conditioned and close to infeasible.

\begin{figure}
\centering
\includegraphics[width=.9\columnwidth]{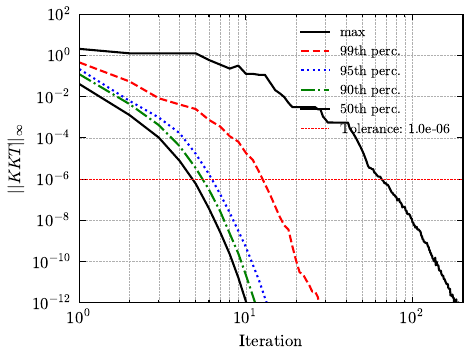}
\caption{Convergence of KKT residuals \eqref{eq:kkt_conditions} measured in the infinity norm over solver iterations for the nonlinear MPC problem on a test dataset of $N_{\text{test}}=5000$ parameter instances $\mathbf{p}^{i}$.
The predictor network is used to initialize the solver, which then refines the solution iteratively.
For each iteration $k$, the figure shows percentiles of the KKT residual infinity norm across all test instances: 50th percentile (median), 90th, 95th, 99th percentiles, and the maximum value.
A tolerance of $\textrm{1e-6}$ is used to determine convergence.
}
\label{fig:kkt_convergence_solver}
\end{figure}

We evaluated the various approaches on a test dataset with $N_{\text{test}} = 5000$ data points, generated in the same closed-loop manner as the training dataset for the approximate MPC.
The convergence behavior of the learning-based solver with predictor is shown in Fig.~\ref{fig:kkt_convergence_solver} for all parameters $\mathbf{p}^{i}$ from the test dataset, i.e., on data points for which a feasible solution exists.
Here, the infinity norm of the residuals of the unmodified KKT conditions \eqref{eq:kkt_conditions} over the iterations $k$ is shown.
The solver iterations are considered converged when the residuals are smaller than $\textrm{1e-6}$.

After fewer than 7 iterations, over 95\% of the test data have already converged, demonstrating that the solver found a solution very quickly in most cases.
Furthermore, convergence was achieved in over 99\% of cases within 20 iterations.
Only in fewer than 1\% of cases, more iterations were necessary to reach the tolerance, which occurs particularly for data points where the problem is ill-conditioned and close to infeasible.

For NMPC applications, the accuracy of the predicted control action $\mathbf{u}_0$ is of particular importance.
Table~\ref{tab:nldi_performance} shows the accuracy of the predicted control action $\mathbf{u}_0$ of the learned solver with predictor compared to IPOPT solutions, approximate MPC, and the predictor without solver.
The accuracy is measured as the absolute deviation of the predicted control action $\mathbf{u}_0$ from the optimal control action $\mathbf{u}_0^{*}$ computed with IPOPT.
To evaluate the suitability of the learned solver for NMPC applications, we also considered the feasibility of the predicted control actions $\mathbf{u}_0$ and the resulting states $\mathbf{x}_1$, which are obtained by evaluating the system dynamics~\eqref{eq:cs_nldi_model} based on the given initial state $\mathbf{x}_0$ and the predicted control action.
Table~\ref{tab:nldi_performance} shows the fraction of test data for which the predicted control action $\mathbf{u}_0$ and the resulting state $\mathbf{x}_1$ violate the state and input constraints by more than $\textrm{1e-6}$.

\begingroup
\setlength{\tabcolsep}{6pt} %
\renewcommand{\arraystretch}{1.2} %
\begin{table*}[t]
    \centering
    \begin{threeparttable}
        \caption{Prediction accuracy comparison for control action $\mathbf{u}_0$ across different methods on the NMPC test dataset ($N_{\text{test}} = 5000$).}
        \label{tab:nldi_performance}
        \centering
        \begin{tabular*}{\tblwidth}{@{}LLC|CCC|CC@{}} %
\toprule
 & & $||KKT||_{\infty}$ & \multicolumn{3}{c|}{$|u_0 - u_0^{*}|$} & $u_0$ viol.\tnote{a} & $x_1$ viol.\tnote{a} \\
\midrule
  &  & max & median & 95th perc. & max & fraction & fraction \\
\midrule
Approx. MPC &  & N/A & 4.39e-03 & 7.94e-03 & 4.50e-02 & 2.23e-01 & 0.0 \\
\cline{1-8}
Predictor &  & 2.09e+00 & 7.95e-03 & 3.33e-02 & 1.12e-01 & 2.43e-01 & 0.0 \\
\cline{1-8}
\multirow[t]{4}{*}{Solver w. Predictor}\tnote{b} & k=1 & 1.27e+00 & 3.37e-04 & 2.56e-03 & 2.42e-02 & 2.71e-01 & 0.0 \\
 & k=5 & 6.02e-01 & 7.84e-08 & 4.76e-06 & 8.52e-03 & 2.20e-02 & 0.0 \\
 & k=20 & 3.13e-03 & 6.39e-08 & 2.22e-06 & 1.28e-03 & 1.20e-03 & 0.0 \\
 & k=65 & 9.08e-07 & 6.30e-08 & 2.11e-06 & 1.28e-03 & 0.0 & 0.0 \\
\cline{1-8}
\bottomrule
\end{tabular*}
        
        \begin{tablenotes}[flushleft]
            \item Shows absolute deviations $|\mathbf{u}_0 - \mathbf{u}_0^{*}|$ from IPOPT optimal solutions and corresponding KKT residual infinity norms.
            \item [a] This indicates the fraction of instances violating constraints by more than $\textrm{1e-6}$, obtained by applying the predicted control action to the system \eqref{eq:cs_nldi_model}.
            \item [b] Results show solver performance after different numbers of iterations $k$, demonstrating how accuracy improves with additional solver refinement steps.
        \end{tablenotes}
    \end{threeparttable}
\end{table*}
\endgroup

Here, it can be observed that better accuracies compared to approximate MPC are already achieved after a single solver step.
Furthermore, the solver converges after a maximum of 65 iterations, and the error in the control actions vanishes. 
The solver with predictor shows no violation of the considered constraints when executed until convergence, while even with a smaller number of iterations, significantly better constraint satisfaction is achieved compared to approximate MPC.
A small error in the predicted control compared to the IPOPT solutions remains in the worst case, which can be attributed to very flat local minima, as the KKT residuals are below the tolerance of $\textrm{1e-6}$ and no constraint violations occur.

\begin{figure}
\centering
\includegraphics[width=.9\columnwidth]{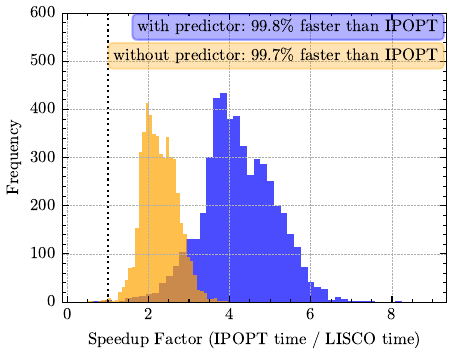}
\caption{Histograms of speedup factors achieved by LISCO compared to IPOPT for the NMPC problem ($N_{\text{test}} = 5000$). The figure shows two distributions: LISCO with predictor (blue) and LISCO without predictor (orange). The histograms display the distribution of runtime ratios (IPOPT time / LISCO time), where values greater than 1 indicate LISCO is faster than IPOPT. A tolerance of $\textrm{1e-6}$ on the KKT residual infinity norm is used for all methods. The percentage of runs where LISCO is faster than IPOPT is indicated in the legend.}
\label{fig:speedup_histogram_comparison}
\end{figure}

To evaluate the suitability of the learned solver as a replacement for a classical optimizer like IPOPT, we compared the runtime against IPOPT and further investigated the influence of using the predictor on the solver runtime.
Both the IPOPT optimizer solutions for the test dataset and the solutions of the learned solver were determined on an AMD Ryzen 9 9950X CPU.
\rev{Training is performed offline once and is not included in the reported solve times.}
While evaluating the learned solver on GPU enables significant parallelization, the evaluation of individual data points, as they occur in closed-loop NMPC applications, is slightly faster on the CPU as the evaluation of the nonlinear KKT residuals is more efficient on the CPU.
\del{The results are summarized in Fig.~\ref{fig:speedup_histogram_comparison}.}
\rev{The results are summarized in Fig.~\ref{fig:speedup_histogram_comparison} and Table~\ref{tab:nldi_runtimes}.}

\begingroup
\setlength{\tabcolsep}{6pt} %
\renewcommand{\arraystretch}{1.2} %
\begin{table*}[t]
    \centering
    \begin{threeparttable}
    \caption{\rev{Runtime performance of the LISCO solver (with and without predictor), IPOPT, approximate MPC, and the predictor alone on the NMPC test dataset ($N_{\text{test}} = 5000$).
    Success is defined as satisfying a KKT residual infinity norm below $\textrm{1e-6}$.
    Approximate MPC and the predictor do not reach this tolerance.
    Entries marked with $-$ are not applicable.}}
    \label{tab:nldi_runtimes}
    \centering
    \begin{tabular*}{\tblwidth}{@{}LCCCCCC}
\toprule
 &  & Solver w. Predictor & Solver w/o. Predictor & IPOPT & Approx. MPC & Predictor \\
\midrule
\multirow[t]{4}{*}{Speedup over IPOPT} & max & 8.91 & 4.55 & - & 140.05 & 121.11 \\
 & 95th perc. & 5.69 & 3.01 & - & 83.24 & 71.99 \\
 & median & 4.16 & 2.24 & - & 61.47 & 53.16 \\
 & min & 0.60 & 0.32 & - & 41.60 & 35.97 \\
\cline{1-7}
\multirow[t]{4}{*}{Solver Iterations} & max & 57\tnote{a} & 117 & - & - & - \\
 & 95th perc. & 6 & 11 & - & - & - \\
 & median & 5 & 9 & - & - & - \\
 & min & 5 & 8 & - & - & - \\
\cline{1-7}
\multirow[t]{4}{*}{Total Solve Time\tnote{b} [s]} & max & 1.20e-02 & 2.51e-02 & 1.26e-02 & 1.37e-04 & 1.95e-04 \\
 & 95th perc. & 2.05e-03 & 3.15e-03 & 7.47e-03 & 9.20e-05 & 1.14e-04 \\
 & median & 1.21e-03 & 2.37e-03 & 5.52e-03 & 8.86e-05 & 1.10e-04 \\
 & min & 1.13e-03 & 2.01e-03 & 3.74e-03 & 8.04e-05 & 8.31e-05 \\
\cline{1-7}
Success Rate &  & 1.00 & 1.00 & 1.00 & 0.00 & 0.00 \\
\cline{1-7}
\bottomrule
\end{tabular*}
        
    \begin{tablenotes}[flushleft]
        \item [a] \rev{The maximum number of iterations for the CPU evaluation is slightly smaller than in the GPU evaluation due to very minor differences in the line search. Using the 2-norm for the tracking condition in the CPU implementation makes the line search slightly more efficient.}
        \item [b] \rev{The reported total solve times include all online costs at inference, i.e., input and output scaling, neural network evaluation, KKT residual computation, and line-search at each iteration.}
    \end{tablenotes}
    \end{threeparttable}
\end{table*}
\endgroup

The learned solver with predictor demonstrates a significant speedup over IPOPT.
\del{In particular, the median speedup is approximately $4.2\times$ and the maximum speedup reaches $8.9\times$.}
Only in the worst case, where the backtracking line-search is frequently activated due to ill-conditioning, is the solver slightly slower than IPOPT.
This backtracking line-search is a heuristic and could be further improved in future work.
\del{Compared to the solver without predictor, the predictor-based solver is significantly faster.}
\rev{Both LISCO variants achieve a success rate of 1.0, solving all instances to a tolerance of $\textrm{1e-6}$.
The predictor roughly halves the required solver iterations (median 5 vs.\ 9) and nearly doubles the median speedup over IPOPT ($4.2\times$, max $8.9\times$, vs.\ $2.2\times$ without predictor), demonstrating its strong impact on solver performance.
LISCO with predictor achieves a median total solve time of $1.21\,\text{ms}$ compared to $5.52\,\text{ms}$ for IPOPT.}
These results highlight the efficiency of the learned solver architecture for solving parametric nonlinear optimization problems and further demonstrate the benefits of using a predictor to initialize the solver.

\rev{To investigate the scaling of LISCO, particularly with respect to the prediction horizon $N$, the solver with predictor was trained for $N \in \{10, 20, 30\}$, with the network width scaled linearly with $N$ (512, 1024, 1536 neurons per hidden layer) and the number of training epochs as well (500k, 1.0M, 1.5M).
Besides the number of neurons per layer and number of training epochs, the same training hyperparameters as previously described were used for all three values of $N$.
The results are summarized in Table \ref{tab:nldi_scaling_results}.
The median iteration count remains the same (5 iterations), the 99th-percentile iteration count stays similar (10-12 iterations) and all runs converge for all three values of $N$ to the desired tolerance of $\textrm{1e-6}$.
The total solve time of LISCO increases more rapidly with $N$ than that of IPOPT ($1.21\,\text{ms}$ to $5.21\,\text{ms}$ versus $5.52\,\text{ms}$ to $8.74\,\text{ms}$ at the median), because the dense network forward pass scales with the network width while IPOPT, using MUMPS, exploits the block-banded sparsity of the NMPC KKT system.
Consequently, the median speedup over IPOPT decreases from $4.16\times$ at $N=10$ to $1.53\times$ at $N=30$.
In the worst case at $N=30$, the speedup drops below one (min $0.31\times$), corresponding to rare instances in which the backtracking line-search is repeatedly activated.
These results motivate future work on more efficient neural network architectures that can explicitly account for the sparsity of the NMPC problem.}

\begingroup
\setlength{\tabcolsep}{6pt} %
\renewcommand{\arraystretch}{1.2} %
\begin{table}[t]
    \centering
    \begin{threeparttable}
    \caption{\rev{Runtime performance of the LISCO solver with predictor for varying prediction horizons $N \in \{10, 20, 30\}$ on NMPC test datasets ($N_{\text{test}} = 5000$).
    The state and input dimensions ($n_x=2$, $n_u=1$) are not changed, while the number of neurons per layer and the number of training epochs are scaled according to the increased problem size.
    Success is defined as satisfying a KKT residual infinity norm below $\textrm{1e-6}$.}}
    \label{tab:nldi_scaling_results}
    \centering
    \begin{tabular}{lccc}
\toprule
 & N=10 & N=20 & N=30 \\
 &  &  &  \\
\midrule
Neurons per Layer & 512 & 1024 & 1536 \\
$n_z$ & 114 & 224 & 334 \\
Training Epochs & 500000 & 1000000 & 1500000 \\
\midrule
Speedup over IPOPT &  &  &  \\
\quad min & 0.60 & 0.80 & 0.31 \\
\quad 1st perc. & 1.99 & 1.62 & 1.03 \\
\quad median & 4.16 & 2.65 & 1.53 \\
\quad max & 8.91 & 4.56 & 2.92 \\
\midrule
Solver Iterations &  &  &  \\
\quad median & 5.0 & 5.0 & 5.0 \\
\quad 99th perc. & 12.0 & 10.0 & 12.0 \\
\quad max & 57.0 & 48.0 & 64.0 \\
\midrule
Total Solve Time [s] &  &  &  \\
\quad median & 1.21e-03 & 2.41e-03 & 5.21e-03 \\
\quad 99th perc. & 3.11e-03 & 4.63e-03 & 1.04e-02 \\
\quad max & 1.20e-02 & 1.66e-02 & 5.35e-02 \\
\midrule
Success Rate & 1.00 & 1.00 & 1.00 \\
\midrule
IPOPT Solve Time [s] &  &  &  \\
\quad median & 5.52e-03 & 6.67e-03 & 8.74e-03 \\
\quad max & 1.26e-02 & 2.06e-02 & 1.69e-02 \\
\bottomrule
\end{tabular}

    \end{threeparttable}
\end{table}
\endgroup

\subsection{Nonconvex Parametric Problem}
Next, we evaluate LISCO on a nonconvex problem with significantly more parameters to demonstrate its applicability to larger problems and compare its performance to the state-of-the-art learning-based approaches DC3 \cite{dontiDC3LearningMethod2021} and PDL \cite{parkSelfSupervisedPrimalDualLearning2023}.
For this purpose, we applied the nonconvex quadratic program (QP) formulation that was previously also used for the evaluation of DC3 and PDL.
The optimization problem is formulated as follows:
\begin{equation}\label{eq:nonconvex_qp}
    \begin{aligned}
        \min_{\mathbf{w}} \quad & \frac{1}{2} \mathbf{w}^{\top}\mathbf{Q}\mathbf{w} + \mathbf{c}^{\top}\sin(\mathbf{w})\\
        \mathrm{s.t.} \quad & \mathbf{A}\mathbf{w} = \mathbf{p},\\
        & \mathbf{G}\mathbf{w} \leq \mathbf{h},
    \end{aligned}
\end{equation}
where $\mathbf{w} \in \mathbb{R}^{100}$ are the decision variables to be optimized, $\mathbf{p} \in \mathbb{R}^{50}$ are the parameters, $\mathbf{Q} \in \mathbb{R}^{100 \times 100}$ is a strictly positive definite matrix, $\mathbf{A} \in \mathbb{R}^{50 \times 100}$ is a matrix for the equality constraints, and $\mathbf{G} \in \mathbb{R}^{50 \times 100}$ is a matrix for the inequality constraints.
The matrices $\mathbf{Q}$, $\mathbf{A}$, and $\mathbf{G}$, as well as the vectors $\mathbf{h}$ and $\mathbf{c}$, were generated as proposed by \cite{dontiDC3LearningMethod2021} to enable direct comparability with results from the literature.
Furthermore, this instantiation ensures that the problems are always feasible as long as the parameters $\mathbf{p}$ are sampled from the interval $[-1,1]^{50}$.
The task of the learned solver is to determine a solution $\mathbf{z}^*$ for a given problem instance with fixed values for $\mathbf{Q}$, $\mathbf{A}$, $\mathbf{G}$, $\mathbf{h}$, and $\mathbf{c}$, given the respective problem parameters $\mathbf{p}$.
Due to the dimension of the parameter vector $\mathbf{p}$ of $n_p=50$, this is a challenging task that cannot be satisfactorily solved with pure supervised learning approaches, similar to approximate MPC, or requires very large amounts of training data.

This optimization problem is nonconvex since the objective function in \eqref{eq:nonconvex_qp} contains the nonlinear sine term.
Therefore, the convexification strategy \eqref{eq:convexification} must also be applied in this case for computing the training loss to ensure that the assumptions of Lemma \ref{lemma:loss_function_per_sample} are satisfied.
Here, the modified Hessian matrix for the convexified formulation was based on the positive definite quadratic matrix of the objective function: $\mathbf{H} = \mathbf{Q}$.

To ensure comparability with results from the literature, we used the same problem instantiation routine as in \cite{dontiDC3LearningMethod2021}, which was also used for generating the results of PDL \cite{parkSelfSupervisedPrimalDualLearning2023}.
Additionally, we generated 5 independent problem instances by creating the matrices $\mathbf{Q}$, $\mathbf{A}$, $\mathbf{G}$ and the vectors $\mathbf{h}$, $\mathbf{c}$ with different random initializations.
The parameters $\mathbf{p}$ were uniformly sampled from the interval $[-1,1]^{50}$ and for each of the 5 problem instances, $N_{\text{test}} = 1000$ test data points were generated.

Similar to the previous NMPC example, we first trained a neural network predictor using randomly sampled problem parameters $\mathbf{p}^{i}$ from the interval $[-1,1]^{50}$.
Subsequently, a neural network solver was trained using the previously trained predictor to generate initial iterates via Algorithm~\ref{alg:solver_training}.
\rev{Furthermore, another solver was trained without relying on the predictor, using random initial iterates instead.}
For both predictor and solver training, a smoothing and penalization parameter of $\epsilon = \textrm{1e-8}$ and $\rho = 0.8$ were used in the Fischer-Burmeister function \eqref{eq:background_fb_for_nlp}.
For solver training, a tolerance of $\delta = \textrm{1e-16}$ for the squared 2-norm in the loss function \eqref{eq:loss_function_per_sample} and a maximum number of iterations of $N_{\text{max}} = 5000$ were applied.
A problem-specific scaling factor of $\gamma = 0.01$ was used in \eqref{eq:solver_step}.
Similar to the NMPC example, individual data points were resampled after $N_{\text{resample}} = 5$ iterations without improvement, the KKT residual norm was clipped to $1.0$ during training, and regularization with $\omega = \textrm{1e-2}$ was applied.
The hyperparameters used for training the predictor and solver are summarized in Table~\ref{tab:hyperparameters_nonconvex_qp}.
To ensure comparability of the predictor network with results from the literature, we used a neural network with two hidden layers of 512 neurons each.
For both networks, the GeLU activation function \cite{hendrycksGaussianErrorLinear2016} was used for the hidden layers, along with a linear activation in the output layer.
The training was implemented in PyTorch \cite{paszkePyTorchImperativeStyle2019} and performed on an NVIDIA RTX 4090 GPU.
Adam \cite{kingmaAdamMethodStochastic2017} with the amsgrad variant \cite{reddiConvergenceAdam2019} was used for optimization in all cases.

\begin{table}[t]
    \caption{Hyperparameters and training times for the predictor and solvers on the nonconvex QP problem instances. The training was performed for 5 independent problem instances and the training times are averaged.}
    \label{tab:hyperparameters_nonconvex_qp}
    \begin{threeparttable}
        \centering
        \begin{tabular*}{\tblwidth}{@{} LCC@{} }
\toprule
 & Predictor & Solver w. Predictor \\
\midrule
Hidden Layers & 2 & 2 \\
Neurons per Layer & 512 & 2048 \\
Batch Size & 4096 & 4096 \\
Epochs & 50k & 500k \\
Learning Rate & 1e-03 & 1e-03 \\
Avg. Training Time [min] & 1.78 & 72.54 (74.32)\tnote{a} \\
\bottomrule
\end{tabular*}

        \begin{tablenotes}[flushleft]
            \item [a] The predictor training time is added in parentheses.
        \end{tablenotes}
    \end{threeparttable}
\end{table}

\begin{figure}
\centering
\includegraphics[width=.9\columnwidth]{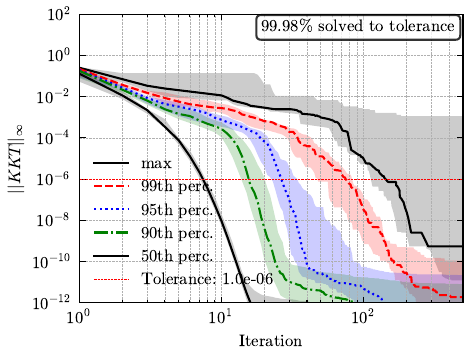}
\caption{Convergence of KKT residuals \eqref{eq:kkt_conditions} measured in the infinity norm over solver iterations for the nonconvex QP problem, aggregated across 5 independent problem instances with $N_{\text{test}}=1000$ parameter instances $\mathbf{p}^{i}$ each.
The predictor network is used to initialize the solver, which then refines the solution iteratively.
For each iteration $k$, the solid lines show the median values across the 5 problem instances for different percentiles of the KKT residual infinity norm: 50th percentile (median), 90th, 95th, 99th percentiles, and the maximum value.
The shaded areas indicate the range from minimum to maximum values across the 5 problem instances.
A tolerance of $\textrm{1e-6}$ is used to determine convergence.}
\label{fig:kkt_convergence_nonconvex_qp}
\end{figure}

The convergence behavior of the learning-based solver with predictor for the 5 different problem instantiations is shown in Fig.~\ref{fig:kkt_convergence_nonconvex_qp} for all parameters $\mathbf{p}^{i}$ from the test dataset.
The solver tolerance for convergence on the infinity norm of the unmodified KKT residuals \eqref{eq:kkt_conditions} was set to $\textrm{1e-6}$.

\rev{The figure reports the median, 90th, 95th, and 99th percentile KKT residuals as well as the range from minimum to maximum values across the five independent problem instances.}
It can be observed that the solver converges in over 95\% of the cases after fewer than 30 iterations to the specified tolerance.
Furthermore, convergence was achieved in all but one case of the 5000 total test data points across the 5 problem instances.

Table \ref{tab:nonconvex_qp_results} presents the optimality gaps and constraint violations of the different approaches, where the results are averaged over 5 independent problem instances and the standard deviations are given in parentheses.
The results for PDL and DC3 are taken from the work of \cite{parkSelfSupervisedPrimalDualLearning2023}, where the same routines for problem instantiation were used and the results were also averaged over 5 independent problem instances.
It can be observed that LISCO with predictor achieves significantly better optimality gaps than PDL and DC3, while the predictor alone already shows moderate performance that is comparable to PDL and only exhibits higher maximum constraint violations.
\rev{This improvement amounts to several orders of magnitude in terms of optimality gap.}

\begingroup
\setlength{\tabcolsep}{6pt} %
\renewcommand{\arraystretch}{1.2} %
\begin{table*}[t]
    \centering
    \begin{threeparttable}
        
        \caption{Performance comparison of different methods on the nonconvex QP problem with 50 parameters and 100 decision variables.}
        \label{tab:nonconvex_qp_results}

\begin{tabular*}{\tblwidth}{@{}LLCCCCCC@{}} %
\toprule
Method &  & Max. Opt. Gap(\%) & Opt. Gap(\%) & Max. Eq. & Max. Ineq. & Mean Eq. & Mean Ineq. \\
\midrule
Predictor &   & 0.479 (0.139) & 0.189 (0.084) & 0.032 (0.005) & 0.059 (0.033) & 0.006 (0.001) & 0.000 (0.000) \\
\cline{1-8}
\multirow[t]{3}{*}{Solver w. Predictor} & k=1 & 0.103 (0.088) & 0.005 (0.001) & 0.009 (0.007) & 0.023 (0.010) & 0.000 (0.000) & 0.000 (0.000) \\
 & k=10 & 0.056 (0.109) & 0.000 (0.000) & 0.004 (0.008) & 0.001 (0.001) & 0.000 (0.000) & 0.000 (0.000) \\
 & final iter. & 0.000 (0.000) & 0.000 (0.000) & 0.000 (0.000) & 0.000 (0.000) & 0.000 (0.000) & 0.000 (0.000) \\
\cmidrule{1-8} %
PDL & & - & 0.324 (0.051) & 0.004 (0.001) & 0.001 (0.000) & 0.001 (0.000) & 0.000 (0.000)\\
DC3 & & - & 4.103 (0.151) & 0.000 (0.000) & 0.000 (0.000) & 0.000 (0.000) & 0.000 (0.000)\\
\bottomrule
\end{tabular*}

        \begin{tablenotes}[flushleft]
            \item The results are averaged over 5 independent problem instances, with standard deviations in parentheses and $N_{\text{test}} = 1000$ test data points for each instance.
            \item The optimality gap is defined as the relative difference between the cost of the predicted solution and the optimal solution found by IPOPT: $(q(\mathbf{w})-q(\mathbf{w}^{*}))/|q(\mathbf{w}^{*})|$.
            \item The constraint violations are measured as the maximum violation of equality and inequality constraints.
            \item The results for PDL and DC3 are taken from \cite{parkSelfSupervisedPrimalDualLearning2023}, where the same instantiation routines were used and also averaged over 5 independent problem instances.
        \end{tablenotes}
    \end{threeparttable}
\end{table*}
\endgroup

\begin{figure}
\centering
\includegraphics[width=.9\columnwidth]{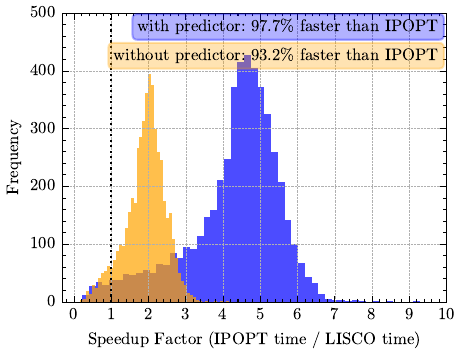}
\caption{\del{Histogram of speedup factors achieved by LISCO compared to IPOPT for the nonconvex QP problem, aggregated across 5 independent problem instances with $N_{\text{test}}=1000$ parameter instances each. The histogram shows the distribution of runtime ratios (IPOPT time / LISCO time), where values greater than 1 indicate LISCO is faster than IPOPT. A tolerance of $\textrm{1e-6}$ on the KKT residual infinity norm is used for both methods. The percentage of runs where LISCO is faster than IPOPT is indicated in the legend.}\rev{Histograms of speedup factors achieved by LISCO compared to IPOPT for the nonconvex QP problem, aggregated across 5 independent problem instances with $N_{\text{test}}=1000$ parameter instances each. The histograms show the distribution of runtime ratios (IPOPT time / LISCO time), where values greater than 1 indicate LISCO is faster than IPOPT. A tolerance of $\textrm{1e-6}$ on the KKT residual infinity norm is used for both methods. The percentage of runs where LISCO is faster than IPOPT is indicated in the legend.}}
\label{fig:speedup_histogram_nonconvex_qp}
\end{figure}

We evaluate the performance of the solver in terms of solve times, comparing it to IPOPT.
\rev{As in the NMPC case study, training is performed offline once and is not included in the reported solve times.}
For this purpose, the solver network was evaluated on the CPU using the same test dataset as before.
\del{The results are summarized in the form of a histogram in Fig.~\ref{fig:speedup_histogram_nonconvex_qp}.}
\rev{The results are summarized in the form of histograms in Fig.~\ref{fig:speedup_histogram_nonconvex_qp} and in Table~\ref{tab:nonconvex_qp_runtimes}.}
The values are presented as aggregated results over all five problem instances.
As can be seen, LISCO is significantly faster than IPOPT on average in reaching the tolerance of $\textrm{1e-6}$, while only the worst-case performance is slightly worse than IPOPT's due to ill-conditioning in some rare cases.
\rev{LISCO with predictor achieves convergence in all but one test instance, achieving a success rate of 0.9998.
The predictor more than halves the required solver iterations (median 7 vs.\ 18) and more than doubles the median speedup over IPOPT ($4.53\times$, vs.\ $1.98\times$ without predictor), demonstrating its strong impact on solver performance.
LISCO with predictor achieves a median total solve time of $4.59\,\text{ms}$ compared to $21.4\,\text{ms}$ for IPOPT.}
\rev{The relative speedup over IPOPT is comparable to that of the NMPC case study despite the higher problem dimension, indicating that the online cost of LISCO scales well with problem size.}

\begingroup
\setlength{\tabcolsep}{6pt} %
\renewcommand{\arraystretch}{1.2} %
\begin{table*}[t]
    \centering
    \begin{threeparttable}
    \caption{\rev{Runtime performance of the LISCO solver (with and without predictor), IPOPT, and the predictor alone on the nonconvex QP problem, aggregated over 5 independent problem instances with $N_{\text{test}} = 1000$ test data points each.
    Entries marked with $-$ are not applicable.}}
    \label{tab:nonconvex_qp_runtimes}
    \centering
    \begin{tabular*}{\tblwidth}{@{}LLCCCC}
\toprule
 &  & Solver w. Predictor & Solver w/o. Predictor & IPOPT & Predictor \\
\midrule
\multirow[t]{4}{*}{Speedup over IPOPT} & max & 9.28 & 4.15 & - & 684.68 \\
 & 95th perc. & 5.89 & 2.70 & - & 482.23 \\
 & median & 4.53 & 1.98 & - & 401.78 \\
 & min & 0.22 & 0.16 & - & 323.06 \\
\cline{1-6}
\multirow[t]{4}{*}{Solver Iterations} & max & 177 & 244 & - & - \\
 & 95th perc. & 25 & 46 & - & - \\
 & median & 7 & 18 & - & - \\
 & min & 6 & 12 & - & - \\
\cline{1-6}
\multirow[t]{4}{*}{Total Solve Time\tnote{a} [s]} & max & 1.14e-01 & 1.38e-01 & 3.64e-02 & 7.91e-05 \\
 & 95th perc. & 1.53e-02 & 2.68e-02 & 2.56e-02 & 5.98e-05 \\
 & median & 4.59e-03 & 1.06e-02 & 2.14e-02 & 5.37e-05 \\
 & min & 3.09e-03 & 6.66e-03 & 1.72e-02 & 4.63e-05 \\
\cline{1-6}
Success Rate &  & 0.9998 & 0.9970 & 1.00 & 0.00 \\
\cline{1-6}
\bottomrule
\end{tabular*}
        
    \begin{tablenotes}[flushleft]
        \item [a] \rev{The reported total solve times include all online costs at inference, i.e., input and output scaling, neural network evaluation, KKT residual computation, and line-search at each iteration.}
    \end{tablenotes}
    \end{threeparttable}
\end{table*}
\endgroup

\rev{
The two case studies on the nonlinear NMPC problem and the nonconvex QP problem establish applicability and good performance of LISCO on the considered problem classes, but do not constitute a universal generalization claim across arbitrary nonconvex problems.
More generally, larger primal-dual dimensions require more network capacity and make accurate update directions harder to learn, although this complexity growth affects classical solvers as well.
Ill-conditioning, as can be measured by the condition number of the modified KKT matrix, further amplifies this difficulty, because small prediction errors translate into weak residual reduction.
The implementation mitigates these effects through input/output scaling of the KKT vector and the predicted step (see Fig.~\ref{fig:solver_zoom_figure}), a training procedure that retains slowly improving iterates and thereby extends exposure to ill-conditioned but feasible cases (see Algorithm~\ref{alg:solver_training}), and a backtracking line search at inference.
For very ill-conditioned instances, additional problem-level scaling may still be required.
Beyond the horizon-scaling study reported in Table~\ref{tab:nldi_scaling_results}, important directions for future work include a more systematic empirical investigation across higher state dimensions and tighter constraint configurations, adaptive step-size strategies, and improved handling of ill-conditioned problems.}

\section{Conclusion}
\label{sec:conclusion}

This work presents a learning-based iterative solver for constrained optimization (LISCO) that addresses fundamental challenges in real-time parametric optimization, specifically in the context of model predictive control (MPC) and other applications requiring fast and accurate solutions to parametric nonlinear optimization problems.

We introduce a novel two-stage architecture consisting of a neural network predictor that generates initial primal-dual estimates, followed by a learned iterative solver that refines these estimates to high accuracy.
Because full primal-dual solutions are available, optimality can be directly certified through KKT error evaluation.
LISCO can be trained in a fully self-supervised manner without requiring pre-solved optimizer solutions, eliminating the dependency on expensive pre-sampled training data.
This is based on a novel loss function that relies on KKT residuals and for which we provide theoretical guarantees that its minima are exclusively at KKT points. This property is a prerequisite for the proposed self-supervised training to be well-posed. %

Furthermore, we introduce a convexification procedure that enables the application of LISCO to nonconvex problems while maintaining the theoretical foundations.
The architecture supports native GPU parallelization using standard machine learning libraries.

The effectiveness of LISCO is demonstrated through two case studies: an NMPC problem for a nonlinear double integrator and a nonconvex parametric optimization problem with 50 parameters, which presents a significant challenge for learning-based approaches due to its high dimensionality.
These studies demonstrate that LISCO achieves significant speedups over classical optimizers like IPOPT while providing superior accuracy compared to alternative learning-based approaches such as approximate MPC, PDL, and DC3.

Future work will consider the derivation of rigorous convergence guarantees for the learned iterative solver.

\section*{Declaration of competing interest}
The authors declare that they have no known competing financial interests or personal relationships that could have appeared to influence the work reported in this paper.

\section*{Data availability}
The code necessary to reproduce the presented results is openly available under \url{https://github.com/lukaslueken/lisco-paper}.

\bibliographystyle{cas-model2-names}

\bibliography{2026_rico}

\end{document}